\documentclass[sigconf]{acmart}

\usepackage{pifont}
\newcommand{\cmark}{\ding{51}}%
\newcommand{\xmark}{\ding{55}}%

\usepackage{microtype}
\usepackage{graphicx}
\usepackage{subfigure}
\usepackage{booktabs} %
\usepackage{natbib}
\usepackage{xcolor}
\usepackage{tikz}
\usepackage{url}
\usepackage{algorithm}
\usepackage{algorithmic}
\usepackage{enumitem}
\usepackage{xspace}
\newcommand\scalethreepics{0.27}
\setcopyright{rightsretained}

\def\xb{{\mathbf x}}

\def\Ccal{\mathcal{C}}

\DeclareMathOperator*{\argmax}{arg\,max}

\newcommand{\blackbox}{model-agnostic \text{ }}

\begin{document}

\copyrightyear{2021}
\acmYear{2021}
\acmConference[AIES '21] {Proceedings of the 2021 AAAI/ACM Conference on AI, Ethics, and Society}{May 19--21, 2021}{Virtual Event, USA.}
\acmBooktitle{Proceedings of the 2021 AAAI/ACM Conference on AI, Ethics, and Society (AIES '21), May 19--21, 2021, Virtual Event, USA}
\acmPrice{}
\acmISBN{978-1-4503-8473-5/21/05}
\acmDOI{10.1145/3461702.3462629}

\fancyhead{}

\title{Fair Bayesian Optimization}

\author{Valerio Perrone}
\affiliation{\institution{Amazon Web Services}}
\email{vperrone@amazon.com}

\author{Michele Donini}
\affiliation{\institution{Amazon Web Services}}
\email{donini@amazon.com}

\author{Muhammad Bilal Zafar}
\affiliation{\institution{Amazon Web Services}}
\email{zafamuh@amazon.com}

\author{Robin Schmucker}
\authornote{Work done as part of an internship at Amazon Web Services.}
\affiliation{\institution{Carnegie Mellon University}}
\email{rschmuck@andrew.cmu.edu}

\author{Krishnaram Kenthapadi}
\affiliation{\institution{Amazon Web Services}}
\email{kenthk@amazon.com}

\author{C\'{e}dric Archambeau}
\affiliation{\institution{Amazon Web Services}}
\email{cedrica@amazon.com}

\renewcommand{\shortauthors}{Perrone, et al.}

\begin{abstract}
Given the increasing importance of machine learning (ML) in our lives, several algorithmic fairness techniques have been proposed to mitigate biases in the outcomes of the ML models. However, most of these techniques are specialized to cater to a single family of ML models and a specific definition of fairness, limiting their adaptibility in practice. We introduce a general constrained Bayesian optimization (BO) framework to optimize the performance of any ML model while enforcing one or multiple fairness constraints. BO is a model-agnostic optimization method that has been successfully applied to automatically tune the hyperparameters of ML models. We apply BO with fairness constraints to a range of popular models, including random forests, gradient boosting, and neural networks, showing that we can obtain accurate and fair solutions by acting solely on the hyperparameters. We also show empirically that our approach is competitive with specialized techniques that enforce model-specific fairness constraints, and outperforms preprocessing methods that learn fair representations of the input data. Moreover, our method can be used in synergy with such specialized fairness techniques to tune their hyperparameters. Finally, we study the relationship between fairness and the hyperparameters selected by BO. We observe a correlation between regularization and unbiased models, explaining why acting on the hyperparameters leads to ML models that generalize well and are fair.
\end{abstract}



\maketitle

\section{Introduction}
\label{sec:intro}

With the increasing use of machine learning (ML) in domains such as financial lending, hiring, criminal justice, and college admissions, there has been a major concern for the potential for ML to unintentionally encode societal biases and result in systematic discrimination~\citep{angwin_2016,barocas2018fairness,bolukbasi_2016,buolamwini2018gender,caliskan_2017}. For example, a classifier that is only tuned to maximize prediction accuracy may unfairly predict a high credit risk for some subgroups of the population applying for a loan. Extensive work has been done to measure and mitigate biases during different stages of the ML life-cycle~\citep{barocas2018fairness}.

Specifically, a number of methods have been proposed in recent years to reduce unfairness in the ML model outcomes. The key design principle behind all of these fairness interventions is to maximize prediction accuracy subject to some constraints on the fairness of the outcomes \textit{while preserving computational tractability}.%
\footnote{Computational tractability here refers to learning or implementing the fairness intervention in a time-efficient manner.}
However, as an artefact of the need to preserve computational tractability, these fairness interventions suffer from one or more of the following drawbacks. The fairness intervention is %
(i) specific to the model class (e.g., linear models only), %
(ii) limited to a specific definition(s) of fairness, 
(iii) limited to a single, binary sensitive feature,
(iv) requires access to sensitive feature information at prediction time, or %
(v) results in a randomized classifier that may generate different prediction for the same input at different times.
These drawbacks limit the practitioners' ability to deploy fair ML solutions for arbitrary machine learning pipelines
corresponding to different practical goals and constraints.\footnote{A pipeline here refers to a ML model (e.g., DNN) with custom input pre-processing and output post-processing steps attached to it.} For instance, as a result of limitation (i), any ensemble or hybrid learning solutions combining different model classes and/or domain knowledge are ruled out. As a drawback of limitation (iv), the corresponding fairness intervention cannot be used when the sensitive feature information is not available at prediction time, while limitation (v) may result in the same model accepting and denying the loan to the same applicant at different times.

In this work, we draw upon the insight that hyperparameter optimization techniques have been used successfully in real-world to optimize the traditional objective of accuracy for arbitrary learning pipelines.  
For instance, several cloud platforms allow users to bring their data along with their own unknown and proprietary model accompanied by any kind of auxiliary pre/post-processing steps; perform model training; and tune the hyperparameters; \textit{all while} treating the pipeline as an opaque box whose internals cannot be modified~\citep{vizier, perrone2020}.\footnote{SigOpt (\url{https://sigopt.com/}); Cloud AutoML (\url{https://www.blog.google/products/google-cloud/cloud-automl-making-ai-accessible-every-business/}); Optuna (\url{https://optuna.org/}); Amazon AMT (\url{https://docs.aws.amazon.com/sagemaker/latest/dg/automatic-model-tuning.html}).}

Motivated by the versatility of hyperparameter tuning, we present a general constrained Bayesian Optimization (BO) framework to tune the performance of ML models with constraints on fairness.  %
In BO, the desired objective is described through a probabilistic model, which not only predicts the best estimates (posterior means), but also uncertainties (posterior variances) for each hyperparameter configuration. The next configuration to try strikes a balance between exploration (sampling where the objective is most uncertain) and exploitation (minimizing the best estimate)~\cite{Srinivas:12}. As a result, BO finds promising hyperparameter configurations more efficiently than random search. Introductions to Bayesian optimization are provided in \cite{Brochu2010} and \cite{Shahriari2016}.

\begin{figure}
\centering
    \includegraphics[width=0.35\textwidth]{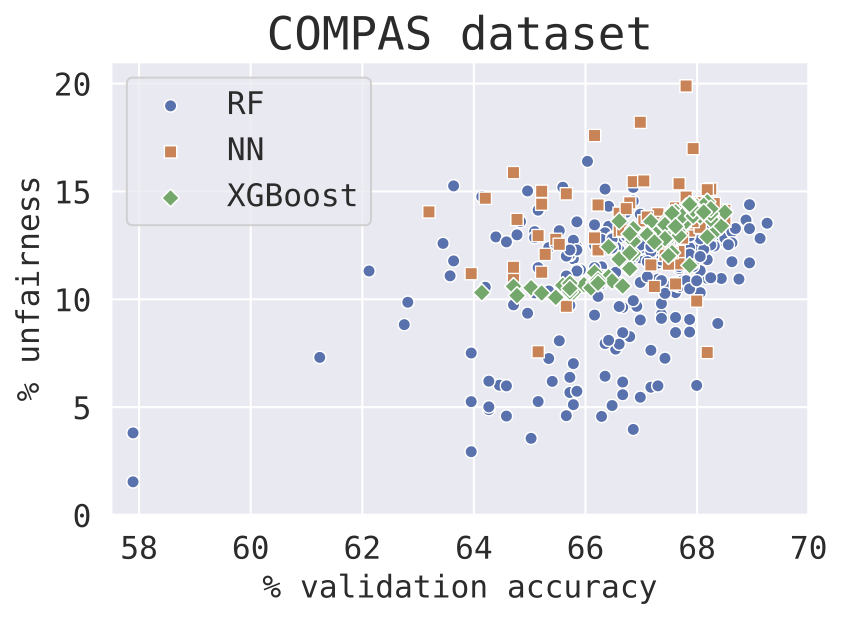}
\caption{Unfairness-accuracy trade-off by varying the hyperparameters of XGBoost, RF, and NN on a recidivism prediction task. Each dot corresponds to a model trained with a different hyperparameter configuration. For a given level of accuracy, models with very different levels of unfairness can be generated simply by changing the model hyperparameters. %
}
\label{fig:compas-intuition}
\end{figure}

Our framework is agnostic to the type of ML model and its training procedure. Hence, it can be applied to arbitrary learning pipelines. 
We demonstrate its effectiveness on several classes of ML models, including random forests, gradient boosting, and neural networks, showing that we can obtain accurate and fair models simply by acting on their hyperparameters. %
 Figure~\ref{fig:compas-intuition} illustrates this idea by plotting the accuracy and unfairness levels achieved by trained gradient boosted tree ensembles (XGBoost~\cite{xgboost}), random forests (RF) and fully-connected feed-forward neural networks (NN), with each dot corresponding to a random hyperparameter configuration. The key observation is that, given a level of accuracy, one can reduce unfairness just by tuning the model hyperparameters. As an example, paying 0.02 points of accuracy (-2.5\%, from 0.69 to 0.67) can give a RF model with 0.08 fewer unfairness points (-70\%, from 0.13 to 0.04). %

It is not uncommon to have laws and regulations enforcing a well-defined fairness metric or ranges of acceptable fairness levels (e.g., the $80\%$ rule for Disparate Impact in \cite{feldman2015certifying}). In these settings, the constrained BO framework we propose is a more suitable choice than multi-objective optimization techniques~\cite{Knowles2006:Parego,Paria2019:Flexible}, which would explore the entire Pareto front of feasible and accurate solutions irrespective of the required fairness level. Further experiments show that our \blackbox approach is also more effective than preprocessing techniques that learn ``fair representations''  of the input data, and is competitive with model-specific methods that have access to the model internals and incorporate fairness constraints as part of the objective during model training.

The paper is organized as follows. Section~\ref{sec:algo_fairness} presents the main ideas of algorithmic fairness, such as state-of-the-art methods and common definitions of fairness. Section~\ref{sec:FairBO} introduces our model-agnostic methodology to optimize model hyperparameters while satisfying fairness constraints. The experimental results in Section~\ref{sec:exp} show that our method compares favourably with state-of-the-art techniques. We also analyze the importance of hyperparameters controlling regularization in the search for fair and accurate models. Finally, Section~\ref{sec:conclusion} presents conclusions and future directions.

We open-sourced our code for constrained BO through AutoGluon~\cite{agtabular}, which is available at the following link: {\url{https://github.com/awslabs/autogluon/}.

\section{Background on Algorithmic Fairness}
\label{sec:algo_fairness}
Algorithmic fairness literature focuses on developing machine learning methods that are both accurate and fair. There is extensive work on identifying and measuring the extent of discrimination (e.g.,~\cite{angwin_2016, caliskan_2017}), and on mitigation approaches in the form of fairness-aware learning methods (e.g.,~\cite{calders2009building, celis2017ranking, donini2018empirical, dwork2012fairness, friedler2016possibility, friedler2018comparative, hardt2016equality, jabbari2017fairness, kamishima2012fairness, woodworth2017learning, zafar2017fairness, zemel2013learning,Zhang2018}).

Today, there is no consensus on a unique definition of fairness and different definitions encode different values that are appropriate in distinct contexts. Moreover, some of the most common definitions are conflicting~\citep{fairness2018verma,kleinberg2018inherent}. Our goal is not to introduce yet another fairness definition, but to propose a flexible methodology that is able to output fair models regardless of the selected criterion we want to enforce. As we will show, in our \blackbox framework we can seamlessly incorporate different definitions, either separately, or simultaneously.

\subsection{Fairness Definitions}

Let $Y$ be the true label (binary),
$S$ the protected (or sensitive) attribute (binary), and $\hat Y$ the predicted label. The most common definitions can be grouped into three categories: (i) considering the predicted outcome given the true label; (ii) considering the true label given predicted outcome; (iii) considering the predicted outcome only. The following are examples of the most used definitions.
\begin{itemize}
\item Equal Opportunity (EO) requires equal True Positive Rates (TPR) across subgroups: $P( \hat{Y} = 1 | Y = 1, S = 0 ) = P( \hat{Y} = 1 | Y = 1, S = 1 )$;
\item  Equalized Odds (EOdd) requires equal False Positive Rates (FPR), in addition to EO; 
\item  Statistical Parity (SP) requires positive predictions to be unaffected by the value of the protected attribute, regardless of the true label $P( \hat{Y} = 1 | S = 0 ) = P( \hat{Y} = 1 | S = 1 )$.
\end{itemize}

Our goal is to find accurate models with a controlled (small) violation of a pre-defined fairness constraint. Hence, following~\cite{donini2018empirical}, we consider the family of $\epsilon$-fair models: a model $\hat{Y}$ is $\epsilon$-fair if it violates the fairness definition by at most $\epsilon \geq 0$.
In the case of EO, a model $\hat{Y}$ is $\epsilon$-fair if the difference in EO (DEO) is at most  $\epsilon$:
    \begin{equation}
    \label{eq-EO-const}
    	|P( \hat{Y} = 1 | Y = 1, S = 0 ) - P( \hat{Y} = 1 | Y = 1, S = 1 )| \leq \epsilon.
    \end{equation}
For EOdd, we have two different constraints simultaneously: the first one is equivalent to DEO and the second one is the difference of FPR (DFP).
Finally, we can similarly define the difference in SP (DSP):
    \begin{equation}
    \label{eq-SP-const}
	|P( \hat{Y} = 1 | S = 0 ) - P( \hat{Y} = 1 | S = 1 )| \leq \epsilon.
    \end{equation}

The inherent trade-offs underlying different notions of fairness has been studied extensively~\cite{dwork2012fairness, friedler2016possibility, kleinberg2018inherent}. Picking the correct definition for the problem at hand is difficult, and it cannot be delegated to an automatic agent in practice. Instead, a human decision is preferred to ensure an informed decision (e.g., with human-in-the-loop approaches \citep{yaghini2019human}).

\subsection{Mitigating unfairness}

Unfairness mitigation techniques can be divided into pre-, in- and post-processing techniques.

Methods in the first family achieve fairness by modifying the data representation, i.e., \textbf{pre-processing} the data, and then to adopt standard ML methods~\cite{calmon2017optimized,zemel2013learning,kamiran2012data}. 
For instance, \cite{zemel2013learning} learn a fair representation by solving an optimization problem with a two-fold goal: encode the data by preserving as much information as possible and obfuscate the sensitive feature group membership.
Whereas techniques like that of~\cite{kamiran2012data} remove the sensitive attribute from the feature set, and rebalance the dataset using SMOTE-like procedures~\citep{chawla2002smote} to under- or over-sample observations from certain sensitive feature groups.
Pre-processing methods are model-agnostic, but their hyperparameters, as well as the ones of the selected ML model still need to be tuned for performance. As we will show shortly, this can negatively impact the fairness and accuracy of the returned solution.

The second family consists of  \textbf{in-processing} methods that enforce model-specific fairness constraints during training (e.g., \cite{donini2018empirical,zafar2017fairness,zafar2019fairness}). These methods are, however, only applicable to a certain model class. For example, the algorithm proposed in \cite{donini2018empirical} can be applied only to kernel machines (such as support vector machines), and only for a single definition of fairness (i.e., EO). %
Although in-processing methods may perform well for the model class that they were designed for, they are often difficult, or sometimes impossible to extend to new model classes. These methods may also introduce new hyperparameters~\cite{zafar2019fairness}, tuning which may require further domain expertise and/or validation.

\textbf{Post-processing} techniques operate by adjusting the decision threshold of a pre-trained classifier to make the eventual outcomes fairer with respect to a given fairness definition. One of the most popular methods in this family is \cite{hardt2016equality}, which also introduces the concept of EO. The main drawback of these techniques is that post-processing the decision threshold is inherently sub-optimal and may lead to worse fairness-accuracy tradeoffs~\cite{woodworth2017learning}. 
Moreover, these techniques are not usable when the sensitive feature information is not available at prediction time, and  might be prone to legal challanges due to the use of sensitive feature for making predictions~\cite{maccarthy2017standards}.

A recently proposed class of methods, referred to as \textbf{meta algorithms}, reduce the fair classification task to a sequence of cost-sensitive classification problems \cite{agarwal2018reductions,agarwal2019fair,kearns18a}.
The solutions to these problems yield a randomized classifier.
As opposed to in-processing methods, the meta algorithms do not depend on the model class of the underlying classifier, rather on the ability to retrain it repeatedly.
The connection between randomized classifiers and fairness (and consequently the connection to differential privacy) has also been studied \citep{oneto2020randomized}.
In the context of empirical risk minimization algorithms, these methods are \blackbox with respect to the ML model, but still need specific implementations based on the selected fairness definition and require to output an ensemble of models.\footnote{Fairlearn code at: \url{https://github.com/fairlearn/fairlearn}.} Similar limitations characterize a number of approaches that either use optimization \citep{kearns18a,Thomas19science} or Bayesian inference \citep{chiappa2019causal,dimitrakakis2019bayesian,FouldsIKP20}. Their implementation has to be tailored to the specific fairness definitions or cannot be applied out-of-the-box to arbitrary ML models and training algorithms. In contrast, we show that our constrained BO approach is agnostic, both, to the selected ML model and to the desired fairness constraint (or ensemble of constraints).

Table~\ref{table:related} summarizes the the existing methods in terms of the capabilities mentioned in Section~\ref{sec:intro}. The table shows that none of the existing methods satisfies all of the desirable practical properties.

Finally, as many existing approaches discussed in this section require tuning certain hyperparameters, we note that our proposal can be used in synergy with these methods to tune their hyperparameters.

\begin{table}[t]
\centering
\small
\begin{tabular}{lccccc}
\toprule
{\bf Method} &  {\bf (i)} &  {\bf (ii)} &  {\bf (iii)} &  {\bf (iv)} & {\bf (v)} \\
\midrule
Zemel et al.~\cite{zemel2013learning} &
\cmark	&	\xmark	&	\xmark	&	\cmark	&	\cmark	\\
Kamiran et al.~\cite{kamiran2012data} &
\cmark	&	\xmark	&	\xmark	&	\cmark	&	\cmark	\\
Donini et al.~\cite{donini2018empirical} &
\xmark	&	\xmark	&	\cmark	&	\cmark	&	\cmark	\\
Zafar et al.~\cite{zafar2019fairness,zafar2017fairness} &
\xmark	&	\xmark	&	\cmark	&	\cmark	&	\cmark	\\
Hardt et al.~\cite{hardt2016equality} &
\cmark	&	\xmark	&	\xmark	&	\xmark	&	\xmark	\\
Agarwal et al.~\cite{agarwal2018reductions} &
\cmark	&	\cmark	&	\cmark	&	\cmark	&	\xmark	\\
Kearns et al.~\cite{kearns18a} &
\cmark	&	\cmark	&	\cmark	&	\cmark	&	\xmark	\\
Ours &
\cmark	&	\cmark	&	\cmark	&	\cmark	&	\cmark \\
\bottomrule
\end{tabular}
\caption{Capabilities of different unfairness mitigation methods w.r.t. desirable practical properties outlined in Section~\ref{sec:intro}: (i) Method caters to arbitrary model classes, (ii) arbitrary fairness definitions, (iii) multiple/polyvalent sensitive features, (iv) can operate without access to sensitive feature at prediction time, and (v) outputs non-randomized predictions.}
\label{table:related}
\end{table}

\section{Fair Bayesian Optimization}
\label{sec:FairBO}
Bayesian optimization (BO) is a well-established methodology to optimize expensive opaque functions (see~\citep{Shahriari2016} for an overview). It relies on a probabilistic model of the unknown target $f(\xb)$ one wishes to optimize. The opaque function $f(\xb)$ is repeatedly queried until one runs out of budget (e.g., time). Queries consist of evaluations of $f$ at hyperparameter configurations $\xb^1,\ldots,\xb^n$ selected according to an explore-exploit trade-off criterion or \emph{acquisition function} \citep{Jones1998}. The hyperparameter configuration corresponding to the best query is then returned. A popular approach is to impose a Gaussian process (GP) prior over $f$ and then compute the posterior GP based on the observed queries $f(\xb^1),\ldots,f(\xb^n)$~\citep{Rasmussen2006}. The posterior GP is characterized by a posterior mean function and a posterior variance function that are required when evaluating the acquisition function for each new query of $f$. 

A widely used acquisition function is the Expected Improvement (EI) \citep{Mockus1978}. This is defined as the expected amount of improvement of an evaluation with respect to the current minimum $f(\xb_{min})$. 
For a Gaussian predictive distribution, EI is defined in closed-form as
\begin{align*}
EI(\xb)&= \mathbf{E}[\max(0,f(\xb_{min})-f(\xb))]  \\
&= \sigma^2(\xb)(z(\xb)\Phi_n(z(\xb))+\phi_n(z(\xb))),
\end{align*}
where $z(\xb) := \frac{\mu(\xb)-f(\xb_{min})}{\sigma^2(\xb)}$. Here, $\mu$ and $\sigma^2$ are respectively the posterior GP mean and variance, and $\Phi_n$ and $\phi_n$ respectively the CDF and PDF of the standard normal. Alternative acquisition functions based on information gain criteria have also been developed~\citep{Hennig2012, Lobato15a, wang17e}. Standard acquisitions focus only on the objective $f(\xb)$ and do not account for additional constraints. In this work, we aim to optimize an opaque function $f(\xb)$ subject to fairness constraints $c(\xb) \leq \epsilon$, with $\epsilon \in \mathbb{R}^+$ determining how strictly the corresponding fairness definition should be enforced.

We next describe Fair Bayesian Optimization (FairBO), our approach to optimize the hyperparameters of an opaque function in a \blackbox manner while satisfying arbitrary fairness constraints (see Algorithm~\ref{alg:constrained-fairness}). For simplicity, we consider the constrained EI (cEI), an established acquisition function to extend BO to the constrained case \citep{Gardner14, Gelbart14, Snoek2015}. Alternatively, FairBO could straightforwardly leverage entropy-based acquisition functions \citep{Lobato15a, Perrone19}. We model the fairness level $c(\xb)$ via an additional GP and weight the EI with the posterior probability of the constraint being satisfied. This gives $cEI(\xb)=EI(\xb) P(c(\xb) \leq \epsilon)$, where the probability of satisfying the fairness constraint $P(c(\xb) \leq \epsilon)$ is computed by evaluating the univariate Gaussian cumulative distribution. In our setting, the feasible hyperparameter configurations are those satisfying the desired fairness constraint (e.g., the DSP across subgroups should be lower than 0.1). The observed feedback on the fairness metric may also be binary in that, rather than the exact fairness value, only a binary signal in $\{-1, +1\}$ may be observed when the constraint is respectively violated or satisfied. In this case, $c(\xb)$ can be modelled through a GP classification model with a Bernoulli likelihood, with the probability of satisfying the constraint being the posterior probability of the positive class~\citep{Rasmussen2006, Gelbart14}.

We define $cEI(\xb)$ as the expected improvement with respect to the current \emph{fair} best, namely the hyperparameter configuration with the best performance metric that satisfies the fairness constraint. As this may not be available in the initial iterations, we start by greedily optimizing $P(c(\xb) \leq \epsilon)$ and then switch to $cEI(\xb)$ (lines 4-10) when the first fair hyperparameter configuration is found. Following the approach from \cite{Gardner14} and \cite{Gelbart14}, this ensures that BO maximizes the probability of finding a feasible configuration before starting to account for performance metrics, which is particularly relevant when the region of feasible hyperparameter configurations is small.
FairBO is straightforward to extend to handle $D$ fairness constraints simultaneously, each with its own upper bound $\{\epsilon_d\}_{d=1}^D$. We merge the fairness constraints into a single binary feedback encoding whether all constraints are satisfied. Assuming independence, one can alternatively place a fairness model $\{ c_d(\xb) \}_{d=1}^D$ on every fairness constraint and  let $P(\xb) = \prod_{d=1}^D  P(c_d(\xb) \leq \epsilon_d)$, each term being the probability of satisfying a fairness constraint.

FairBO assumes the user specifies the desired fairness constraints \emph{a priori}. As a result, the method can explore the region of the hyperparameter space that satisfies the fairness constraints in a data-efficient way. When the user does not specify the desired fairness constraints, one would be forced to rely on a multi-objective strategy to recover the Pareto front of fair and accurate solutions, selecting an acceptable fairness and accuracy level \emph{a posteriori}~\citep{Pfisterer2019, Chakraborty2019, Chakraborty2020}. Multi-objective optimization comes with its own challenges (e.g., how to trade-off multiple objectives). As we will see shortly, it is also inherently more expensive computationally and slower than FairBO to return a solution as it does not focus on the feasible region.

\begin{algorithm}
   \caption{FairBO}
   \textbf{Input}: Initial and total budgets $T_0$, $T$; unfairness bound $\epsilon$; GP prior on objective $f(\xb)$ and fairness $c(\xb)$ model.
   \label{alg:constrained-fairness}
\begin{algorithmic}[1]
   \STATE Evaluate $f(\xb_i)$ and $c(\xb_i)$ for $i=1,...,T_0$ hyperparameters $\xb_i$ from the search space (e.g.,  drawn uniformly at random or from a fixed initial design) and set the used budget $t =T_0$.
   \STATE Define the set of evaluated hyperparameters $\Ccal = \{ (\xb_i, f(\xb_i), c(\xb_i) \}_{i=1}^{T_0}$
    \STATE Compute the posterior GP for the objective and the fairness models based on $\Ccal$.
    \WHILE{ $t < T$}
        \STATE $\xb_{\text{new}} \in  \argmax_{\xb} EI(\xb) P(c(\xb) \leq \epsilon)$.
        \STATE Evaluate $f(\xb_{\text{new}})$ and $c(\xb_{\text{new}})$.
        \STATE Update $\Ccal = \Ccal \cup \{ (\xb_\text{new}, f(\xb_\text{new}), c(\xb_{\text{new}})) \} $
         \STATE Compute the posterior GP for the objective and the fairness models based on $\Ccal$.
        \STATE $t = t + 1$
    \ENDWHILE
    \STATE \textbf{return} Best fair hyperparameter configuration in  $\Ccal$.
\end{algorithmic}
\end{algorithm}

\section{Experiments}
\label{sec:exp}
We demonstrate the benefits of FairBO through extensive experiments. Our evaluation is driven by four key messages. First, we show that FairBO is an efficient strategy to find well-performing and fair hyperparameter configurations, converging to a good solution faster than standard BO and random search. Second, we show that FairBO is agnostic to the fairness constraint and can naturally handle multiple constraints simultaneously. Third, we shed light on the relationship between hyperparameters and fairness metrics, finding a connection between regularization hyperparameters and fairness. Finally, we show that FairBO compares favourably to state-of-the-art model-agnostic and model-specific fairness techniques.

We consider three datasets widely used in the context of fairness: (i) Adult -- Census Income~\citep{Dua:2019}, a binary classification problem with binary gender as sensitive attribute, where the goal is to predict if income exceeds $\$50$K/yr based on census data; (ii) German Credit Data~\citep{Dua:2019}, a binary classification problem with binary gender as sensitive attribute, where the goal is to classify people described by a set of attributes as good or bad credit risks; (iii) COMPAS, a binary classification problem concerning recidivism risk, with binarized ethnic group as sensitive attribute (one group for ``white'' and one for all other ethnic groups).\footnote{COMPAS link: \url{https://github.com/propublica/compas-analysis}. We are aware of the ethics issues concerning COMPAS dataset. We think that recidivism prediction is itself a highly contested area. We decided to use this data only due to its popularity as benchmark dataset in the field, but our goal is not to normalize the definition of social fairness for recidivism prediction as something that can be achieved by only relying on a statistical fairness criterion \emph{as is} without a diligent human intervention in the decision process.} We tune four popular ML algorithms implemented in \texttt{scikit-learn} \citep{pedregosa2011scikit}: XGBoost, Random Forest (RF), a fully-connected neural network (NN), and Linear Learner (LL), optimizing the hyperparameters in Appendix A. We optimize for validation accuracy, with a random 70\%/30\% split into train/validation, and place an upper bound on unfairness (e.g., defined via DSP as per inequality \eqref{eq-SP-const}). All hyperparameter optimization methods are initialized with 5 random hyperparameter configurations. BO and FairBO are implemented in GPyOpt~\citep{gpyopt2016}, with the GP using a Mat\'ern-5/2 covariance kernel with automatic relevance determination hyperparameters, optimized by type-II maximum likelihood \cite{Rasmussen2006}. Results are averaged across 10 repetitions, with 95\% confidence intervals obtained via bootstrapping. All experiments are run on \texttt{AWS} with \texttt{m4.xlarge} machines.

\begin{figure*}[h!]
    \centering
        \includegraphics[width = \scalethreepics\textwidth]{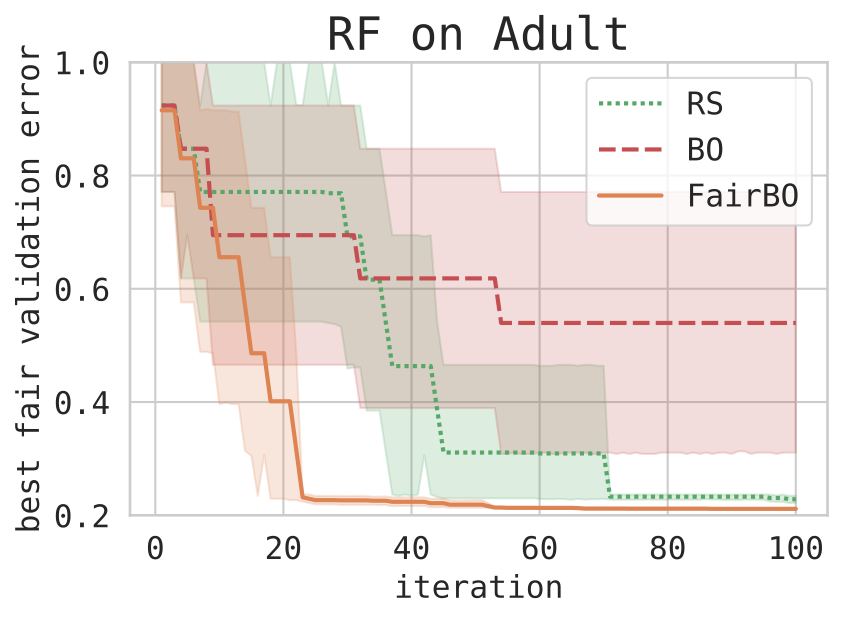}
    \includegraphics[width = \scalethreepics\textwidth]{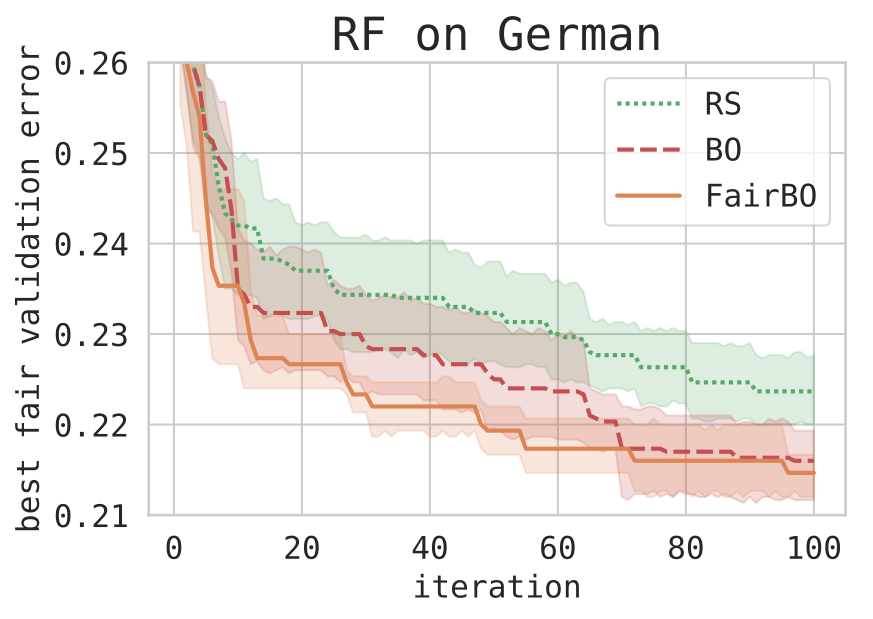}
\includegraphics[width = \scalethreepics\textwidth]{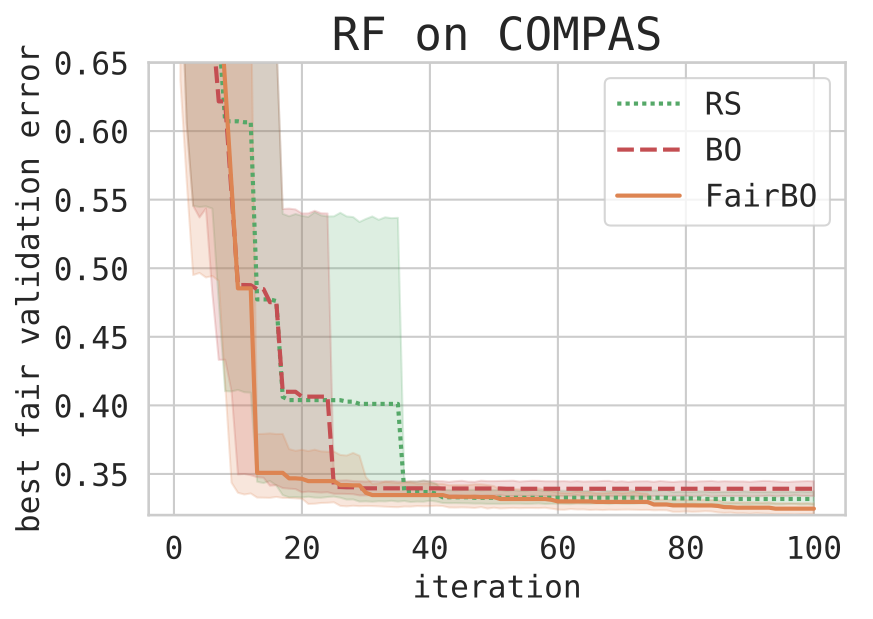}

    \caption{Comparison of RS, BO, and FairBO on the task of tuning RF on Adult, German, and COMPAS. The plot shows the validation error of the best feasible solution found as a function of the number of evaluations. The fairness constraint is DSP $\leq$ 0.05. FairBO finds a more accurate fair model in fewer iterations.}
    \label{fig:bo-performance}
\end{figure*}

\begin{figure*}[h!]
    \centering
    \includegraphics[width = \scalethreepics\textwidth]{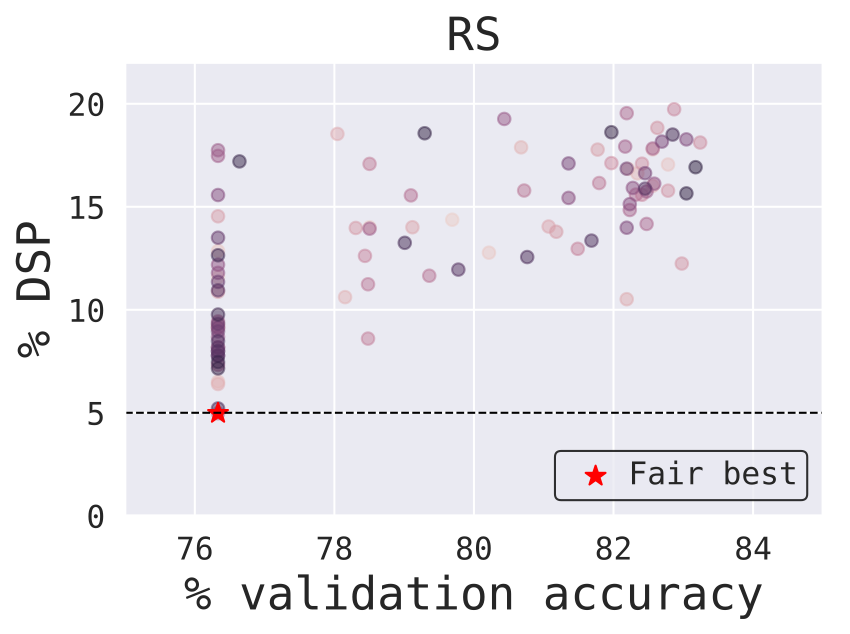}
    \includegraphics[width = \scalethreepics\textwidth]{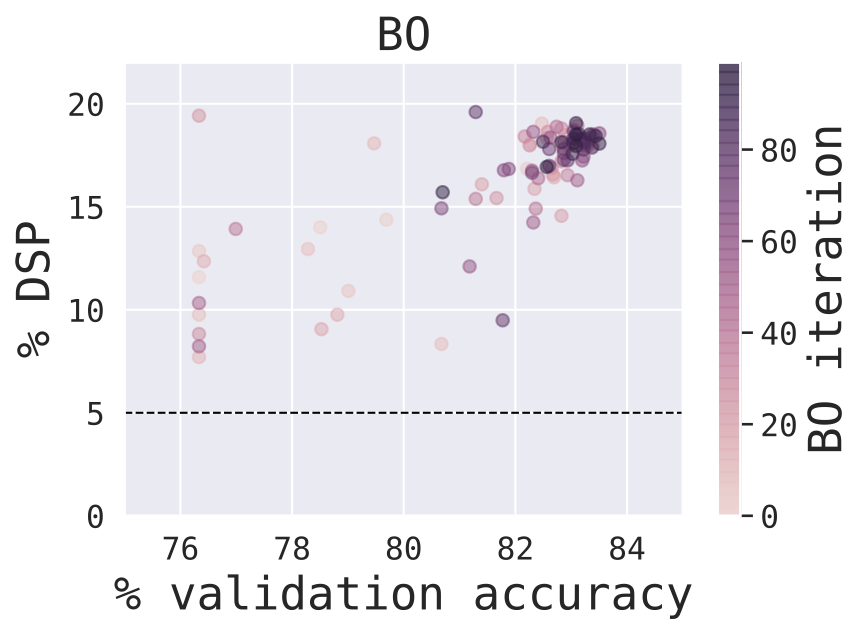}
        \includegraphics[width = \scalethreepics\textwidth]{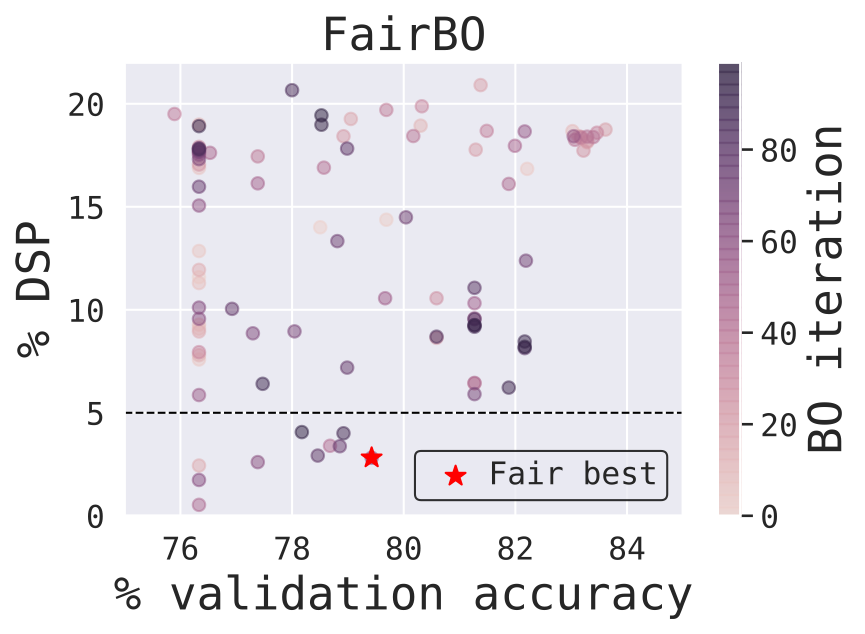}
    \caption{Comparison of RS, BO, and FairBO on the task of tuning RF on Adult. The horizontal line is the fairness constraint, set to DSP $\leq$ 0.05, and darker dots correspond to later BO iterations. Standard BO can get stuck in high-performing yet unfair regions, failing to return a feasible solution (i.e., there is no fair best). RS is more robust than BO, but only finds a fair model with low validation accuracy.}
    \label{fig:adult-results}
\end{figure*}

\subsection{FairBO performance}
We first compare FairBO to Random Search (RS) and standard BO (based on the EI acquisition). Figure \ref{fig:bo-performance} compares the validation error of the fair solution found on Adult, German, and COMPAS with a fairness constraint of DSP $\leq$ 0.05 (i.e., the selected model has a difference in probability of predicting a positive output between the two subgorups that is less or equal than 5\%, see Eq.~\ref{eq-SP-const} with $\epsilon = 0.05$). 
For each seed, all methods are initialized with the same set of five hyperparameter evaluations drawn uniformly at random from the search space. While all methods are eventually expected to find similar solutions with enough budget, we are interested in how fast they find a well-performing fair hyperparameter configuration. As expected, FairBO finds an accurate and fair model more quickly than RS and BO. On Adult, FairBO reaches the fair (local) optimum five times faster than RS. Figure~\ref{fig:adult-results} shows an example run with all tried hyperparameter configurations. Standard BO can get stuck in high-performing yet unfair regions, failing to return a feasible solution. While RS is more robust in that it cannot get stuck exploring unfeasible regions, it only finds a fair solution with the same accuracy of the trivial model always predicting the majority class (i.e., the set of points with accuracy $\approx$ 0.763). In contrast, FairBO is able to sample the fairness-performance space much more effectively. RS samples the space of hyperparameters uniformly, which does not correspond to sampling the fairness-performance space uniformly, while standard BO does not receive any signal from violating the constraint and blindly exploits well-performing yet unfair hyperparameters. 

Analogous results are shown in Appendix B for the problem of tuning XGBoost and NN. We also repeat these experiments with a different instantiation of FairBO through constrained max-value entropy-search (cMES), the state-of-the-art acquisition function for constrained BO based on information gain criteria~\citep{Perrone19}. We show that the two instantiations of FairBO achieve similar performance on these benchmarks. %
Appendix B also includes experiments with a looser fairness constraint DSP $\leq$ 0.15. As expected, the accuracy of the best fair solution is higher under a looser constraint and FairBO still compares favorably to the baselines. In the appendix we also repeat the experiments with a constraint on DEO (inequality \eqref{eq-EO-const}) and confirm the benefits of FairBO, noting that other fairness definitions can be used if desired.

\subsection{Multiple fairness constraints}
In contrast to most algorithmic fairness techniques, FairBO can seamlessly handle multiple fairness definitions simultaneously. We consider 100 iterations of standard BO and FairBO on the problem of tuning RF on Adult and repeat the experiment three times, each time introducing an additional fairness constraint. %
Specifically, we first impose a constraint on DFP, then on both DFP and DEO, and finally on DFP, DEO, and DSP together. All constraint thresholds are set to 0.05 and results are averaged over 10 independent repetitions. Figure~\ref{fig:polar} shows accuracy and three fairness metrics (i.e., one minus unfairness, namely, ($1 - $ DFP), ($1 - $ DEO), ($1 - $ DSP) respectively) of the returned fair solution for RF. Analogous results for XGBoost, NN, and LL are given in Appendix B. Interestingly, FairBO allows us to trade off relatively little accuracy for a fairer solution, which gets progressively fairer as we add more constraints.

\begin{figure*}[h!]
    \centering
    \includegraphics[width = 0.3\textwidth, trim={1.0cm 0.5cm 0.5cm 0.0cm}, clip]{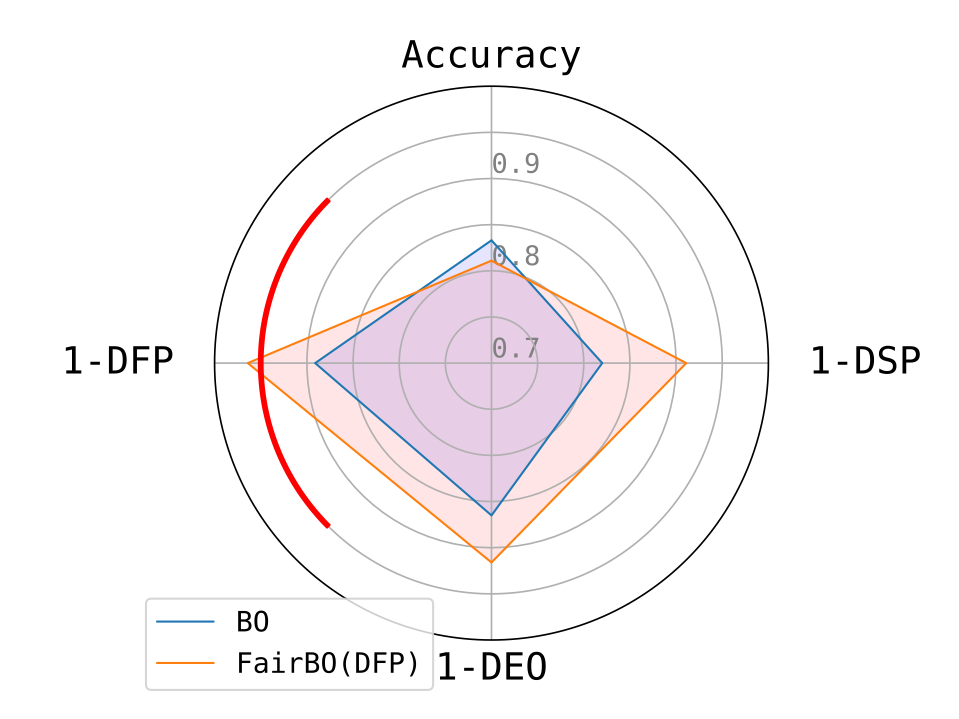}
    \includegraphics[width = 0.315\textwidth, trim={0.2cm 0.5cm 0.5cm 0.0cm}, clip]{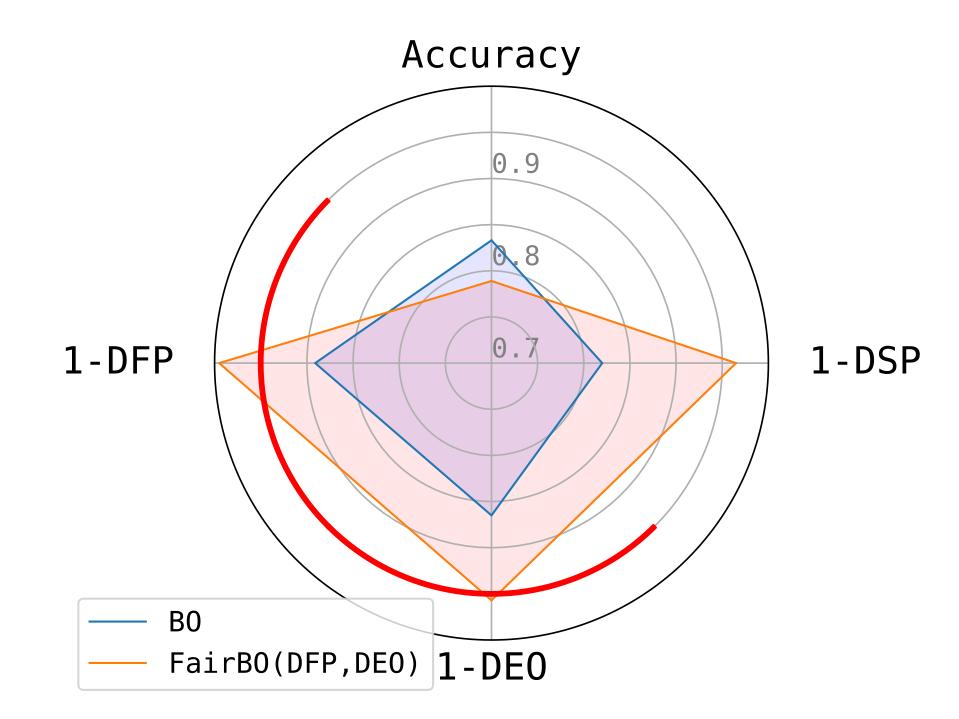}
     \includegraphics[width = 0.315\textwidth, trim={0.2cm 0.5cm 0.5cm 0.0cm}, clip]{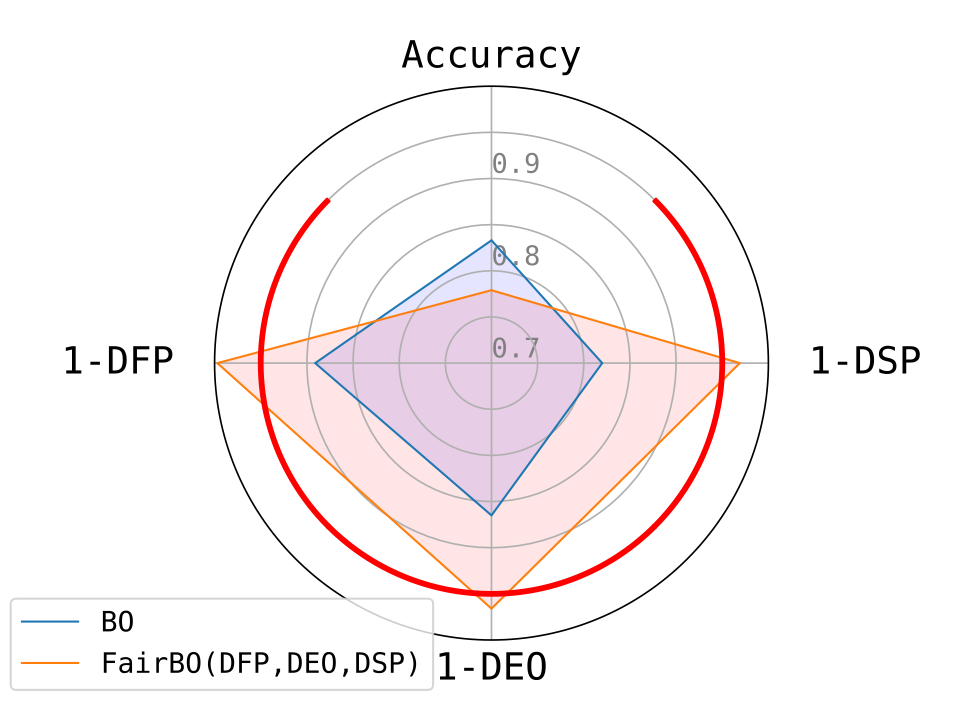}
 \caption{Best RF hyperparameter configuration found on Adult by BO and  FairBO with respectively one, two, and three fairness constraints (left: DFP $\le 0.05$; center: \{DFP $\le 0.05$, DEO $\le 0.05$\}; right: \{DFP $\le 0.05$, DEO $\le 0.05$, DSP $\le 0.05$\}). The red arches represent the enforced level of fairness, i.e. the minimum level of fairness we can accept in order to satisfy the constraints. Compared to BO, FairBO can trade off accuracy for a fairer solution, and can do so with respect to all definitions simultaneously.}
 \label{fig:polar}
\end{figure*}

\subsection{Hyperparameters and fairness}
We showed that hyperparameter tuning can mitigate unfairness effectively. We now investigate more closely the role of each hyperparameter on the unfairness of the resulting model. For each algorithm, we apply fANOVA \citep{Hutter14} to study hyperparameter importance on fairness, defined as DSP (analogous results with DEO are given in Appendix B). Hyperparameter configurations and unfairness metrics are collected from 100 iterations of random search and 10 random seeds, for a total of 1000 data points per algorithm-dataset pair. Figure~\ref{fig:fanova} indicates that the hyperparameters controlling the regularization level tend to have the largest impact on fairness. In the case of RF, the most important hyperparameter is the maximum tree depth; for XGBoost, this is either the L1 weight regularizer \texttt{alpha} or the number of boosting rounds; for NN, Adam's initial learning rate \texttt{eps} plays the biggest role (as we keep the number of epochs fixed, Adam's initial learning rate works indeed as regularizer). Finally, for LL the most relevant hyperparameter influencing fairness is precisely the regularization factor \texttt{alpha}. Figure~\ref{fig:fanova-followup} shows the DSP and accuracy for 100 random hyperparameter configurations for each algorithm, before and after fixing the most relevant hyperparameter detected by fANOVA to the default value from \texttt{scikit-learn}. As expected, freezing these hyperparameters limits the ability of FairBO to provide fair and accurate solutions, leading to fewer fair solutions. %

We conjecture that, by preventing overfitting, the hyperparameters controlling regularization generate models with a lower ability to discriminate among the different values of the sensitive attribute. For example, consider the simple case in which the sensitive feature is uncorrelated with the other features. Assuming we have a linear model $w$, where the entry $w_s \in \mathbb{R}$ is the weight assigned to the sensitive feature, we can bound the DSP as follows:
\begin{align*}
	DSP(w) &= |P_x( w \cdot x \geq 0 | S = 0 ) - P_x( w \cdot x \geq 0 | S = 1 )|\\
	& \leq P_x(-w_s \leq w \cdot x < 0 | S = 0 ).
\end{align*}
The idea is that, given $x$ in subgroup $S=0$ and being $x'$ the same as $x$ where we flipped the value of the sensitive feature to $S=1$, we have $w \cdot x - w \cdot x' = -w_s$. Consequently, a smaller $w_s$ helps obtain a less biased model. 
Indeed, unfairness correlates with the weight assigned to the sensitive feature (or to the sum of the weights assigned to all the features correlated with the sensitive one), and regularization tends to alleviate this. Increasing the regularization in a cross-entropy loss will have a similar effect, giving progressively less data dependent models and steering DSP closer to zero.
Similar insights apply to a more general setting of a non-linear parametric model (e.g., NN) and non-sensitive features: as if the hyperparameter $\texttt{alpha} \to \infty \implies w \to 0$, which in turn means that the  model prediction approaches $0$ regardless of the input features, satisfying all the fairness definitions in Section~\ref{sec:algo_fairness} in the process.

\begin{figure*}[h!]
    \centering
    \includegraphics[width = 0.3\textwidth]{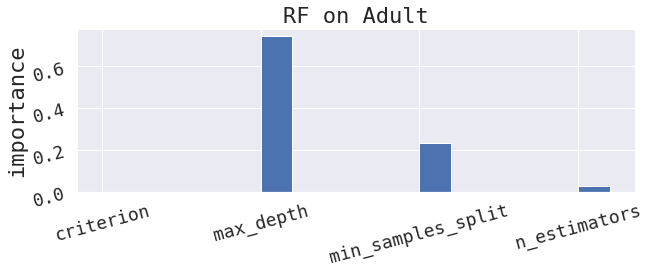}
   \includegraphics[width = 0.3\textwidth]{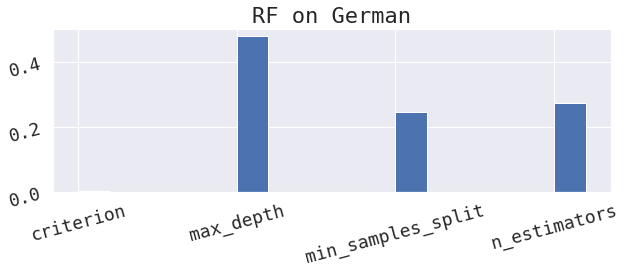}
   \includegraphics[width = 0.3\textwidth]{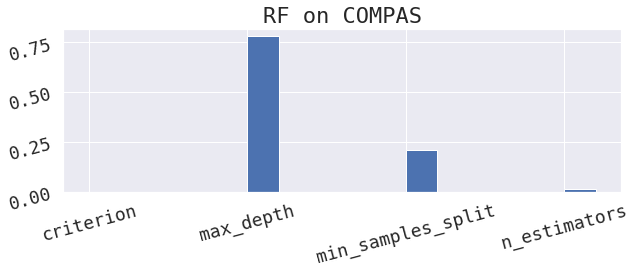}
        \includegraphics[width = 0.3\textwidth]{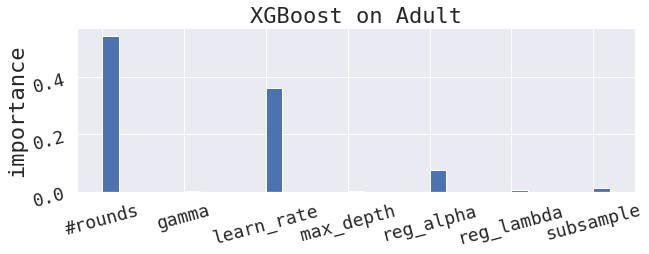}
   \includegraphics[width = 0.3\textwidth]{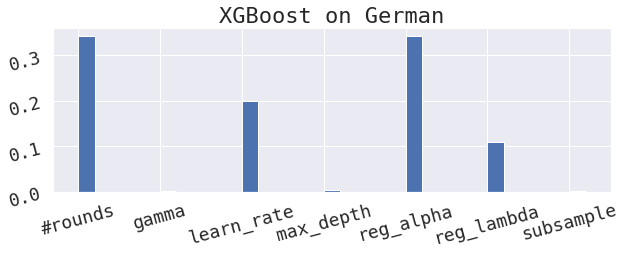}
   \includegraphics[width = 0.3\textwidth]{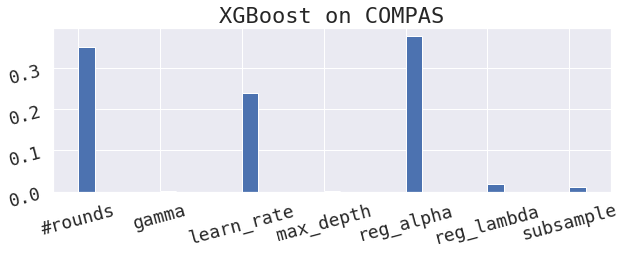}
           \includegraphics[width = 0.3\textwidth]{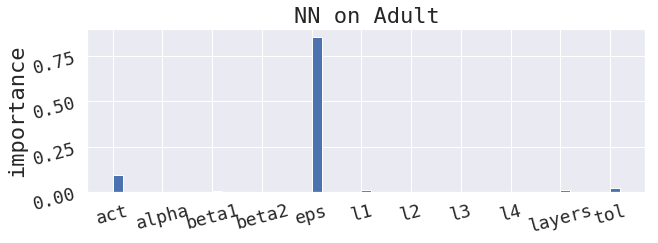}
   \includegraphics[width = 0.3\textwidth]{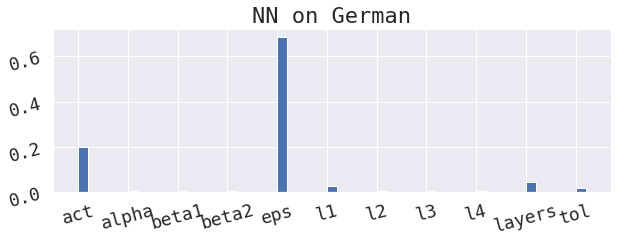}
   \includegraphics[width = 0.3\textwidth]{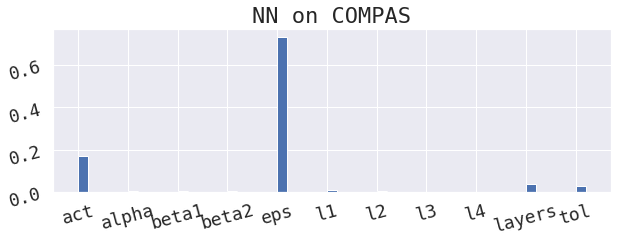}
           \includegraphics[width = 0.3\textwidth]{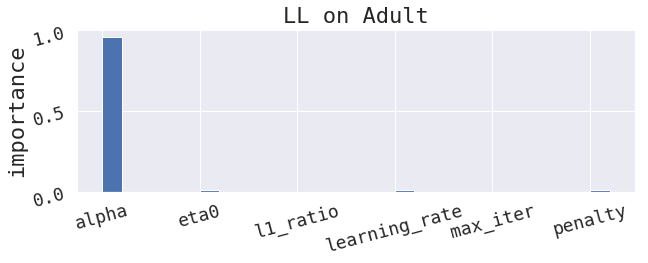}
   \includegraphics[width = 0.3\textwidth]{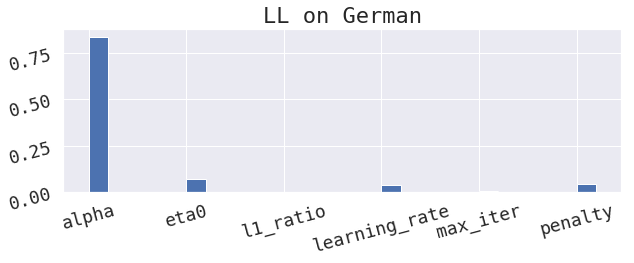}
   \includegraphics[width = 0.3\textwidth]{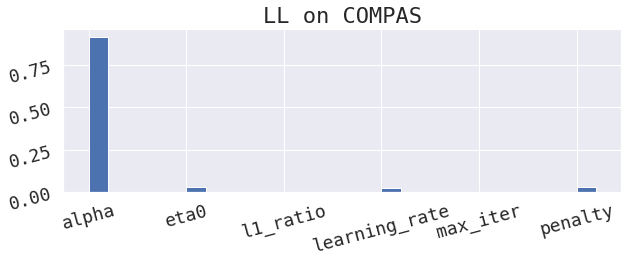}
\caption{Hyperparameter importance on fairness: the role of each tuned hyperparameter on model unfairness (DSP) is evaluated with fANOVA \citep{Hutter14}. Statistics are collected from 100 iterations of random search and 10 seeds. Regularization hyperparameters tend to impact fairness the most (e.g., the most relevant hyperparameter in LL is precisely the regularization factor \texttt{alpha}).}
    \label{fig:fanova}
\end{figure*}

\begin{figure*}[h!]
    \centering
    \includegraphics[width = 0.24\textwidth]{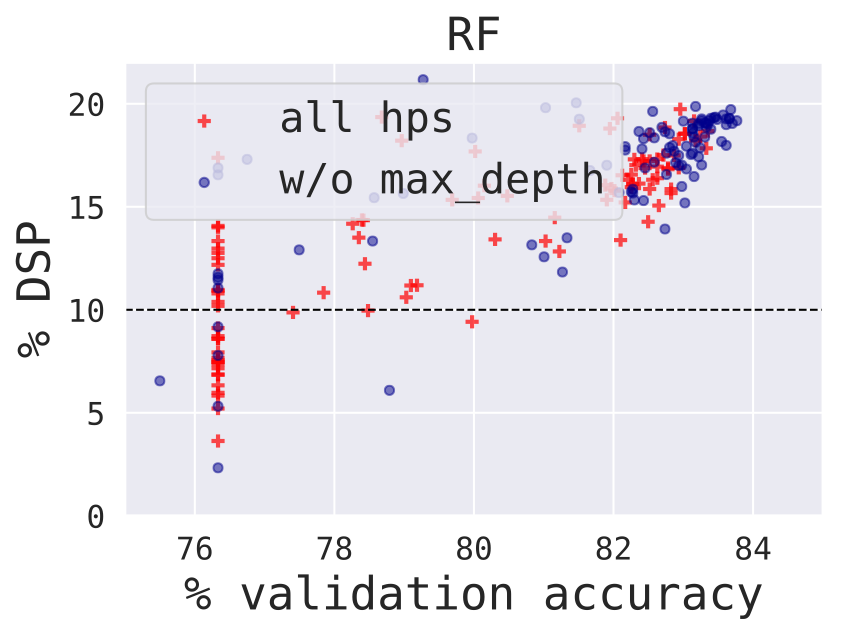}
        \includegraphics[width = 0.24\textwidth]{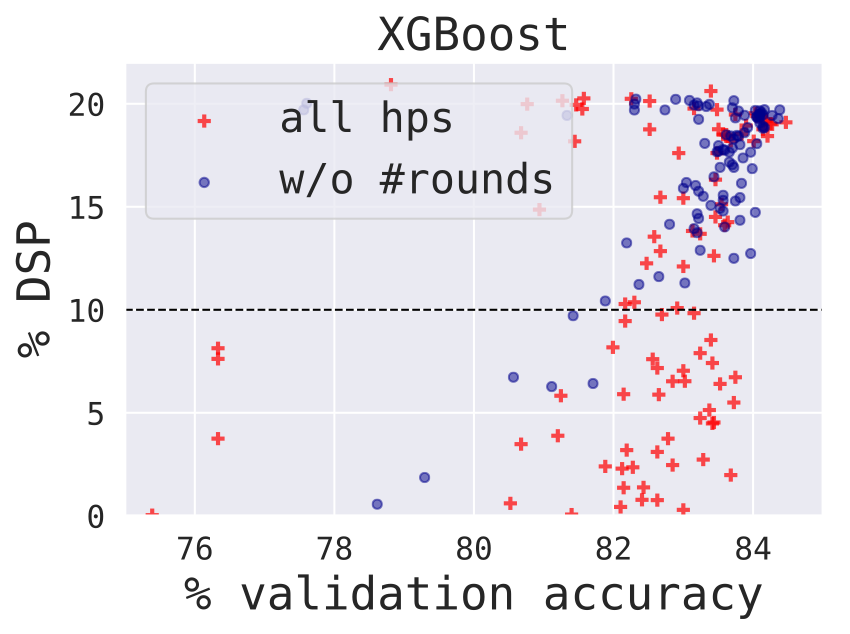}
            \includegraphics[width = 0.24\textwidth]{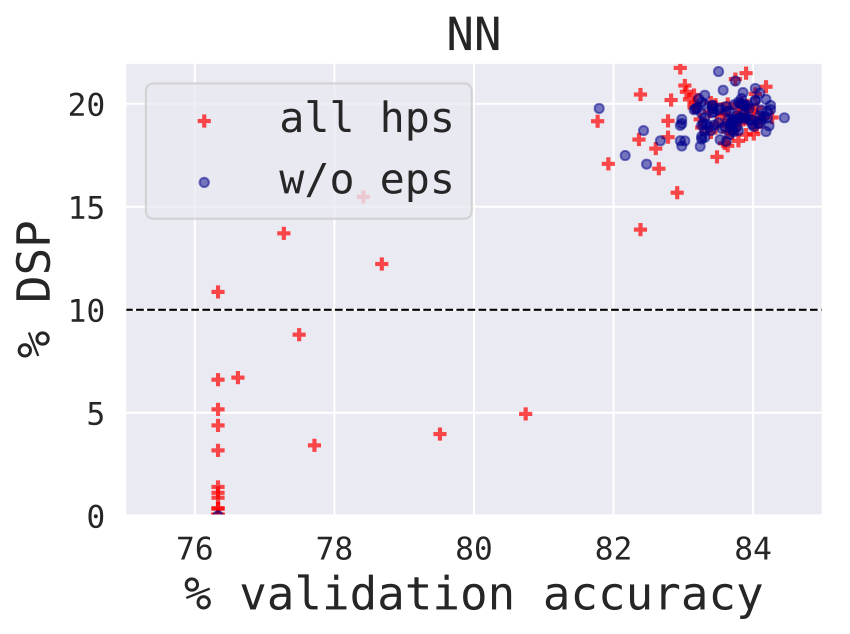}
                \includegraphics[width = 0.24\textwidth]{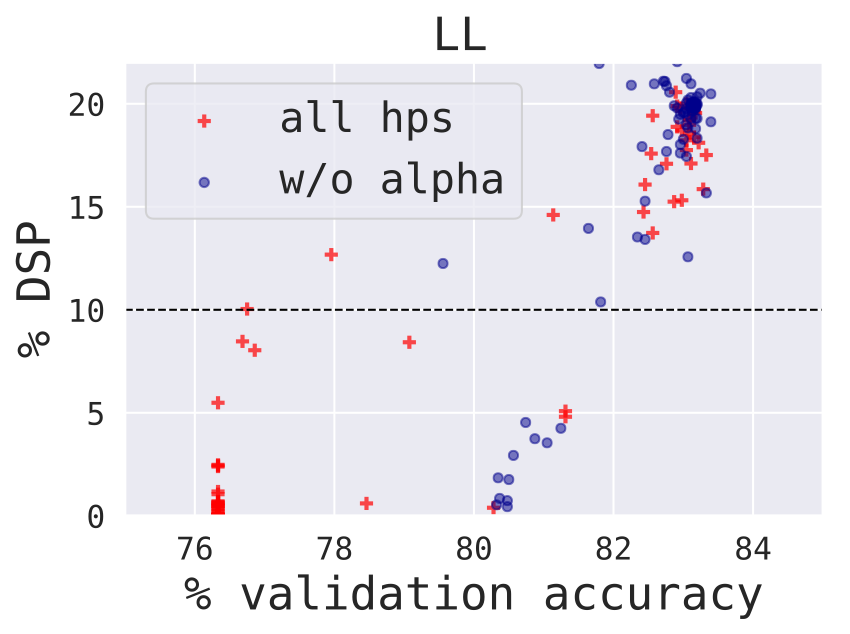}
    \caption{Unfairness vs Accuracy on Adult for 100 random hyperparameter configurations of RF, XGBoost, NN, and LL. For each algorithm, we either vary all hyperparameters or fix the one most correlated with fairness (detected by fANOVA \citep{Hutter14}). As expected, fixing these hyperparameters limits the ability of FairBO to provide fair and accurate solutions, yielding fewer fair configurations (fewer blue dots fall in the feasible area below the horizontal line).}
    \label{fig:fanova-followup}
\end{figure*}

\subsection{Model-agnostic and model-specific techniques}

In the context of algorithmic fairness several \emph{ad-hoc} methods have been proposed. We compare to the method of Zafar \citep{zafar2017fairness}, Adversarial debiasing \citep{Zhang2018}, and Fair Empirical Risk Minimization (FERM)~\citep{donini2018empirical}.\footnote{Code for Zafar from \url{https://github.com/mbilalzafar/fair-classification}; code for Adversarial Debiasing from \url{https://github.com/IBM/AIF360}; code for FERM from \url{https://github.com/jmikko/fair_ERM}.} These methods enforce fairness during training and optimize the parameters of a linear model to make it both accurate and fair with respect to a fixed fairness definition. These methods are not model-agnostic and only apply to linear models. As alternative \blackbox approaches we compare to SMOTE, which preprocesses the data by removing the sensitive feature and rebalancing the observations, as well as FERM preprocessing~\citep{donini2018empirical}, which learns a fair representation of the data before fitting a linear model. We allocate 100 hyperparameter tuning iterations for all approaches.

Table \ref{tab:ferm} shows the best performing fair model found by FairBO on LL compared to the best performing fair model found by each baseline. As expected, FERM achieves higher accuracy, due to the constraint applied directly while training the parameters (as opposed to the hyperparameters) of the linear model. However, the gap in performance with FairBO is modest, and FairBO outperforms both Zafar and Adversarial Debiasing. While conceptually simple, FairBO emerges as a surprisingly competitive baseline that can outperform or compete against these highly specialized techniques. We note that all model-specific techniques tend to find solutions that are fairer than the required constraint. FairBO is also the best model-agnostic method, outperforming both SMOTE and FERM preprocessing. This shows that we can remove bias with a smaller impact on accuracy.

As FairBO only acts on the hyperparameters, it can be used \emph{on top} of model-specific techniques to tune their own hyperparameters. Blindly tuning these hyperparameters can negatively impact the fairness of the resulting solution. We demonstrate this by combining FairBO with Zafar and Adversarial Debiasing, which we found to be sensitive to their hyperparameter settings (unlike FERM). Figure~\ref{fig:zafar-fairhpo} shows that hyperparameter tuning on top of model-specific techniques yields better-performing fair solutions, and FairBO tends to find them more quickly than random search and standard BO. FairBO can be thus used to tune alternative fairness techniques, finding superior fair solutions.

\begin{table*}[h!]
\centering
\caption{Validation error of the best fair models for model-specific (first three rows) and model-agnostic fairness methods. We use the fairness constraint, DSP $\le 0.1$. The best method from each category is highlighted in bold.}
\begin{tabular}{l|ccc}
\hline
Method & Adult & German & COMPAS\\
\hline 
FERM &  \textbf{0.164 $\pm$ 0.010} & \textbf{0.185   $\pm$  0.012} & \textbf{0.285 $\pm$ 0.009} \\
Zafar &  0.187 $\pm$ 0.001 & 0.272 $\pm$ 0.004 & 0.411 $\pm$ 0.063 \\
Adversarial & 0.237 $\pm$ 0.001 & 0.227 $\pm$ 0.008 &  0.327  $\pm$ 0.002 \\
\hline
FERM preprocess &  0.228 $\pm$ 0.013 & 0.231 $\pm$ 0.015 & 0.343 $\pm$ 0.002 \\
SMOTE &  0.178 $\pm$ 0.005 & 0.206 $\pm$ 0.004 & 0.321 $\pm$ 0.002 \\
FairBO (ours) &  \textbf{0.175 $\pm$ 0.007} & \textbf{0.196  $\pm$  0.005} & \textbf{0.307 $\pm$ 0.001} \\
\hline
\end{tabular}
\label{tab:ferm}
\end{table*}

\begin{figure*}[h!]
    \centering
    \includegraphics[width = \scalethreepics\textwidth]{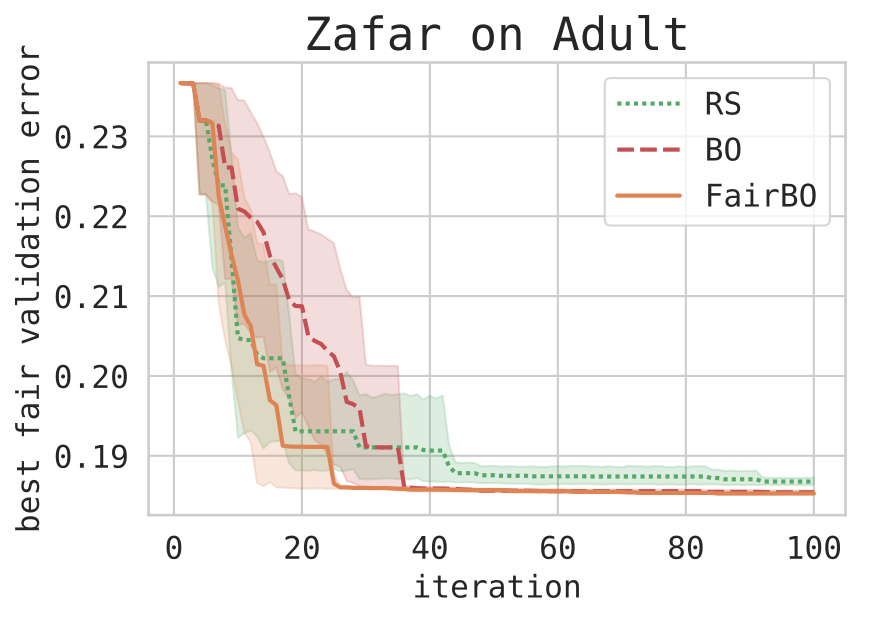}
        \includegraphics[width = \scalethreepics\textwidth]{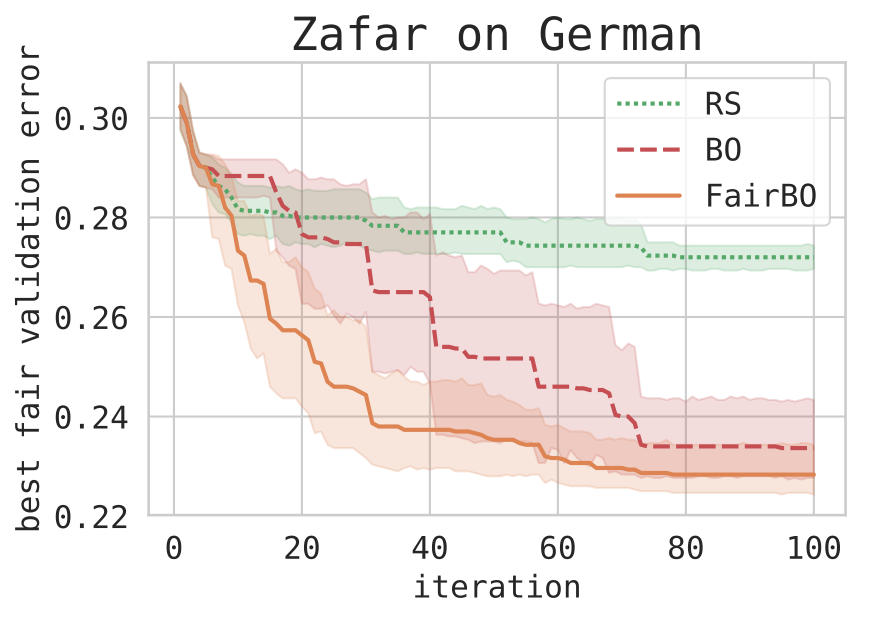}
    \includegraphics[width = \scalethreepics\textwidth]{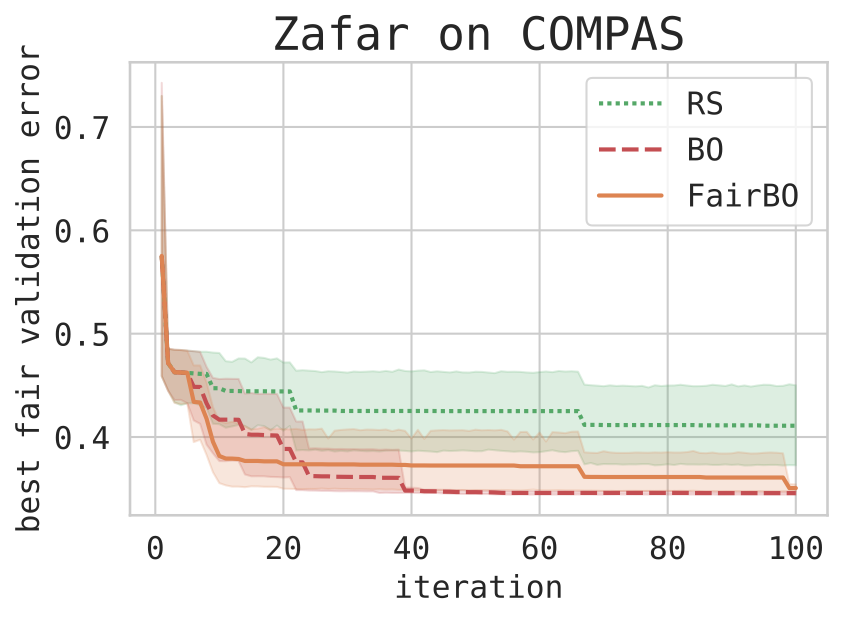}
        \includegraphics[width = \scalethreepics\textwidth]{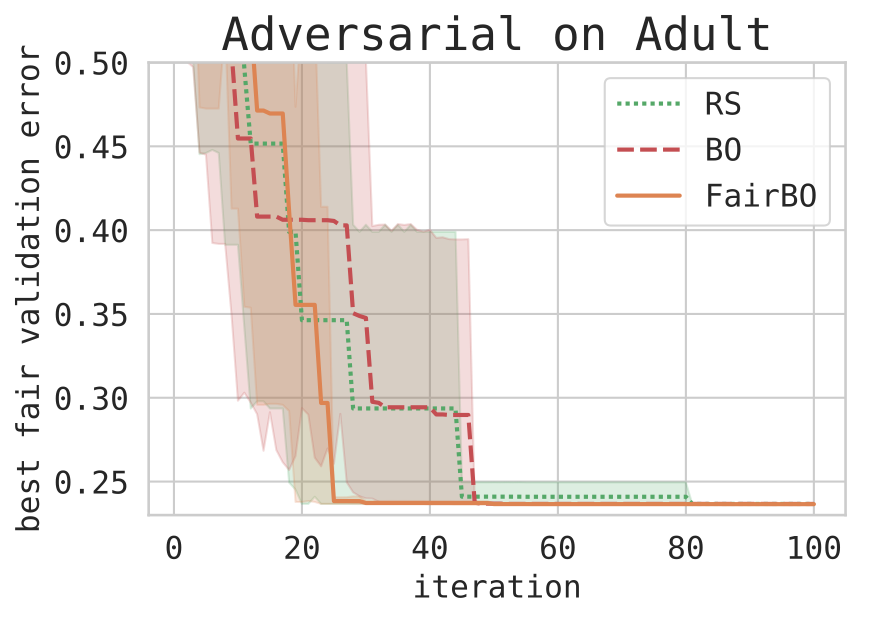}
        \includegraphics[width = \scalethreepics\textwidth]{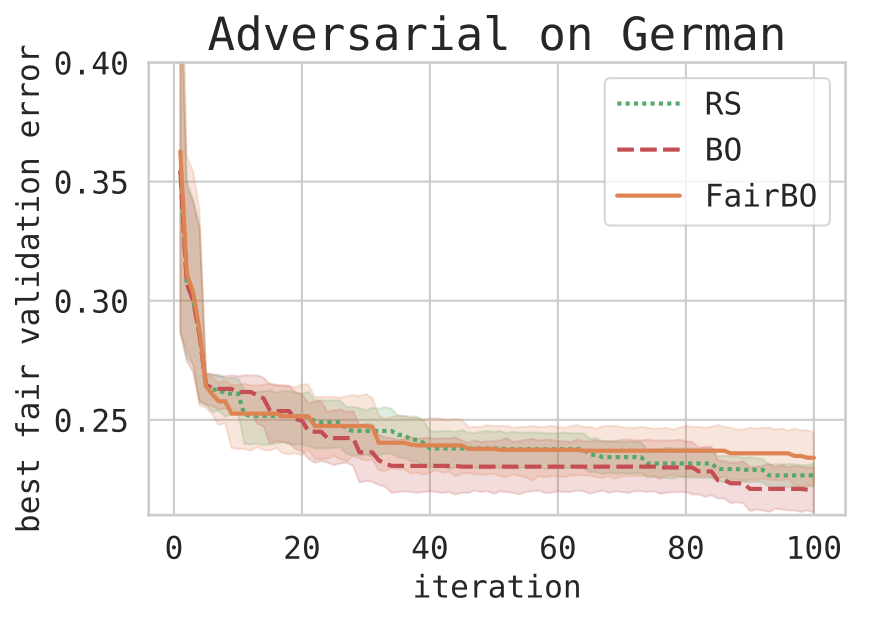}
    \includegraphics[width = \scalethreepics\textwidth]{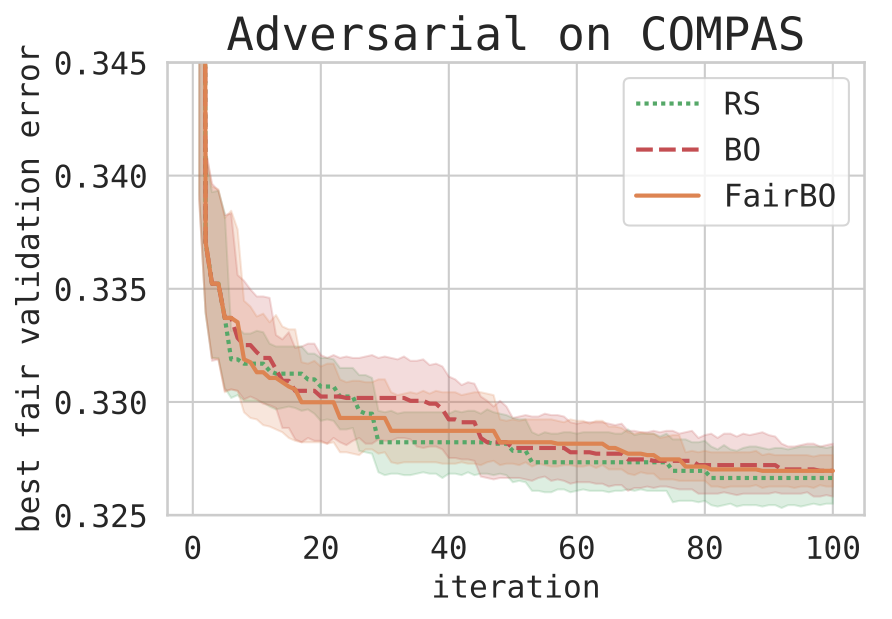}
    \caption{Comparison of RS, BO, and FairBO where Zafar and Adversarial Debiasing are used as base learners. We use the fairness constraint, DSP $\le 0.1$. Hyperparameter tuning on top of model-specific techniques finds better performing fair solutions. The use of FairBO tends to do so in fewer iterations (horizontal axis) than when using RS or BO.}
 \label{fig:zafar-fairhpo}
\end{figure*}

\begin{figure*}[h!]
\centering
    \includegraphics[width= \scalethreepics\textwidth]{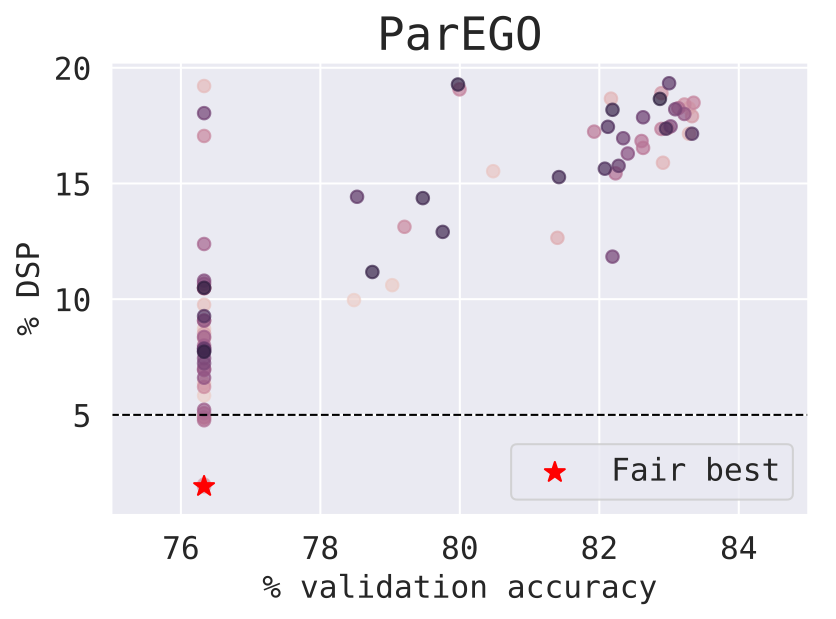}
    \includegraphics[width= \scalethreepics\textwidth]{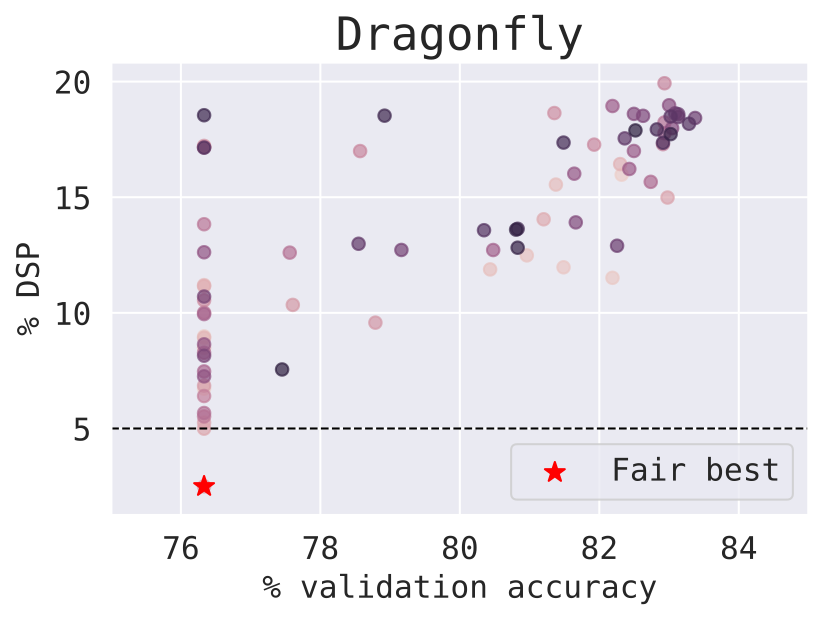}
      \includegraphics[width = \scalethreepics\textwidth]{figures/adult/CEI_statistical_disparity_vs_score_rf_plot_ub005_adult_008.png}
\caption{ParEGO and Dragonfly tuning RF on Adult compared to FairBO. The horizontal line is the fairness constraint (DSP $\leq$ 0.05), and darker dots are later BO iterations. MOBO methods find a better Pareto Front but fail to focus on the feasible region.}
\label{fig:ParEGO}
\end{figure*}

\subsection{Comparison with Multi-objective BO}
Another natural methodology to find a model that is both fair and accurate is to use multi-objective (MO) optimization techniques (e.g., see parallel work in \citep{Cruz2020}). These methods attempt to find the Pareto front corresponding to different trade-offs between the objectives. Hence, they do not require the user to specify the fairness threshold(s) \emph{a priori}. However, as they are trying to find the ensemble of solutions on the Pareto front (as opposed to the single best satisfying the constraint) they incur a much higher computational cost. Depending on the task at hand, it is not rare to have specific regulations and laws imposing a well defined fairness metric and which range of values of the metrics is considered as acceptable -- for example the $80\%$ rule for Disparate Impact (see \cite{feldman2015certifying} for more details). In these situations, FairBO is able to provide a feasible solution faster than MO optimization techniques due to a more focused exploration of the feasible area of the domain.

We compare FairBO with two state-of-the-art MO Bayesian optimization (MOBO): (i) ParEGO~\cite{Knowles2006:Parego}, and (ii) a scalarization based method by Paria, et al.,~\cite{Paria2019:Flexible} implemented in the Dragonfly library~\cite{Kandasamy2020:Tuning}. MOBO aims to find the Pareto front of expensive multi-output opaque functions. %
Figure~\ref{fig:ParEGO} shows the RF models found by MOBO on Adult with the same experimental setup and number of evaluations as Figure~\ref{fig:adult-results}. MOBO methods are outperformed by FairBO in the feasible area (DSP $\leq 0.05$), because the \emph{a priori} knowledge of the constraint enables FairBO to focus its attention on the (possibly small) feasible part of the Pareto front. In addition, FairBO samples the relevant fairness-performance space much more effectively than the multi-objective approaches. Finally, one should note that ParEGO and Dragonfly are designed to fully explore the Pareto front, and thus inevitably scale worse with respect to the number of objectives.

\section{Conclusions}
\label{sec:conclusion}
We showed that tuning model hyperparameters is surprisingly effective to mitigate unfairness in ML and proposed FairBO, a constrained Bayesian optimization framework to jointly tune ML models for accuracy and fairness. FairBO constitutes a strong baseline and is flexible. Indeed, it is model agnostic, can be used with arbitrary fairness definitions, and allows for multiple fairness definitions to be applied simultaneously. The proposed methodology empirically finds more accurate and fair solutions than data-debiasing techniques, while being competitive with the state-of-the-art in algorithm-specific fairness. We also showed that FairBO is preferable over standard BO when tuning the hyperparameters of specialized techniques. Finally, we demonstrated the importance of regularization hyperparameters in yielding fair and accurate models.
Potential directions for future work include applying our framework to regression, image recognition, and natural language processing problems, and covering other fairness definitions, such as with continuous sensitive attributes.

\bibliographystyle{abbrv}
\bibliography{references,bibliography}

\end{document}


\maketitle

\newpage
\appendix
\section{Optimized hyperparameters}
\label{AppendixA}

\subsection{Algorithms}
We considered the problem of tuning four popular ML algorithms, as implemented in \texttt{scikit-learn}: XGBoost (XGB), Random Forest (RF), Neural Network (NN), Linear Learner (LL). In this section, we give more details on the search space over which each hyperparameter was optimized. 

\paragraph{XGBoost}
We consider a 7-dimensional search space:  number of boosting rounds in $\{1, 2, \dots, 256\}$ (log scaled), learning rate in $[0.01, 1.0]$ (log scaled), minimum loss reduction to partition leaf node \texttt{gamma} in $[0.0, 0.1]$, L1 weight regularization \texttt{alpha} in $[10^{-3}, 10^3]$ (log scaled), L2 weight regularization \texttt{lambda} in $[10^{-3}, 10^3]$ (log scaled), subsampling rate in $[0.01, 1.0]$, maximum tree depth in $\{1, 2, \dots, 16\}$.

\paragraph{RF}
We consider a 4-dimensional search space:  number of trees in $\{1, 2, \dots, 64\}$ (log scaled), tree split threshold in $[0.01, 0.5]$ (log scaled), tree maximum depth in $\{1, 2, 3, 4, 5\}$, criterion for quality of split in \{Gini, Entropy\}.

 \paragraph{NN}
We consider an 11-dimensional search space: number of layers in $\{1, 2, 3, 4\}$, each layer size in $\{2, 3, \dots, 32\}$ (log scaled), activation in \{Logistic, Tanh, ReLU\}, tolerance in $[10^{-5}, 10^{-2}]$ (log scaled), L2 regularization in $[10^{-6}, 0.1]$ (log scaled), and Adam parameters: initial learning rate \texttt{eps} in $[10^{-6}, 10^{-2}]$ (log scaled), beta1 and beta2 in $[10^{-3}, 0.99]$ (log scaled).

\paragraph{LL}
We consider a 6-dimensional search space: iteration count in $\{1, 2, \dots, 128\}$, regularization type in \{L1, L2, ElasticNet\}, Elastic Net mixing parameter in $[0, 1]$, regularization factor \texttt{alpha} in $[10^{-3}, 10^3]$ (log scaled), initial learning rate \texttt{eta0} in $[10^{-4}, 0.1]$ (log scaled), learning rate schedule in \{Constant, Optimal, Invscaling, Adaptive\}.

\subsection{Baselines}

In addition to the hyperparameters of the tuned algorithms, we allocated 100 iterations of random search to tune each baseline, namely FERM, Zafar, Adversarial, FERM preprocess and SMOTE.

\paragraph{FERM}
We considered 2 hyperparameters for FERM. The L2 regularization coefficient C in $[0.001, 10.0]$ (log scaled), and FERM's epsilon-fairness threshold in $[0.0, 1.0]$.

\paragraph{Zafar}
We tuned 3 hyperparameters:  L1 regularization coefficient in $[10^{-3}, 10.0]$ (log scaled), L2 regularization coefficient in $[0.001, 10.0]$ (log scaled), and Zafar's epsilon-fairness threshold in $[0.0, 1.0]$.

\paragraph{Adversarial}
We tuned 4 different hyperparameters: the adversary loss weight in $[0.0, 1.0$], the number of epochs for training in $\{1, 2, \dots, 500\}$ (log scaled), the batch size in $\{32, 33, \dots
, 256\}$ (log scaled), and the number of hidden units of the network between in $\{20, 21, \dots, 5000\}$ (log scaled).

\paragraph{FERM preprocess}
FERM preprocessing learns a fair representation of the dataset, which is then fed to LL. Hence, we tuned the same 6 hyperparameters as per the original LL.

\paragraph{SMOTE}
In addition to the 6 hyperparameters of LL, we jointly tuned 2 hyperparameters controlling the degree of dataset rebalancing: oversampling rate of the less frequent class in $[0.3, 1.0]$ and number of neighbors to generate synthetic examples in  $\{1, 2, \dots, 20\}$.

\section{Additional Experiments}
\label{AppendixB}
Compared to Random Search (RS) and standard BO, FairBO explores the fair regions of the hyperparameter space more quickly and tends to return a more accurate, fair solution. This appendix presents additional experiments with different fairness thresholds, definitions and algorithms. We also study the impact of each hyperparameter on the deviation from equal opportunity (DEO) of the resulting model.

\begin{figure*}[h]
    \centering
        \includegraphics[width = 0.32\textwidth]{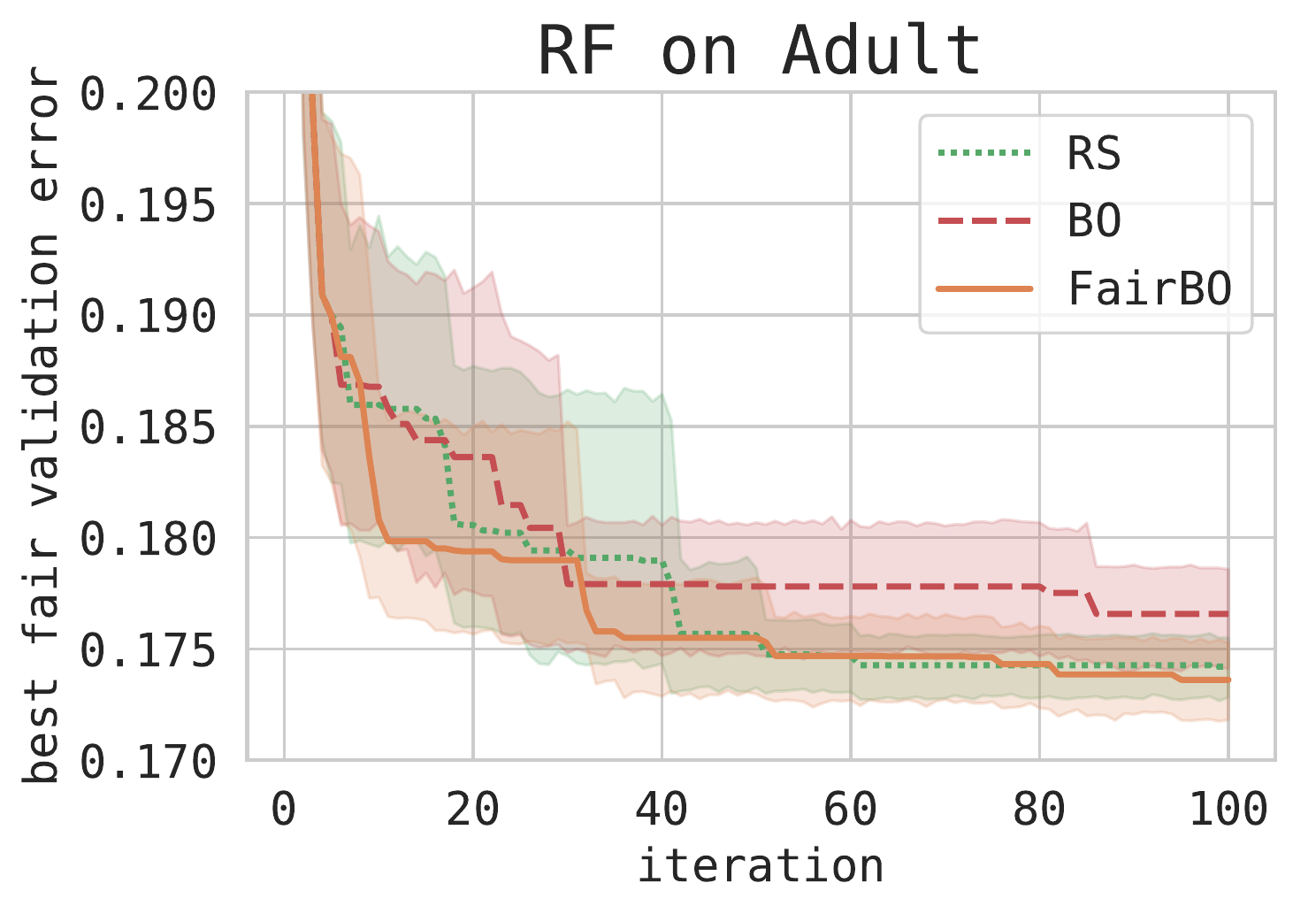}
    \includegraphics[width = 0.32\textwidth]{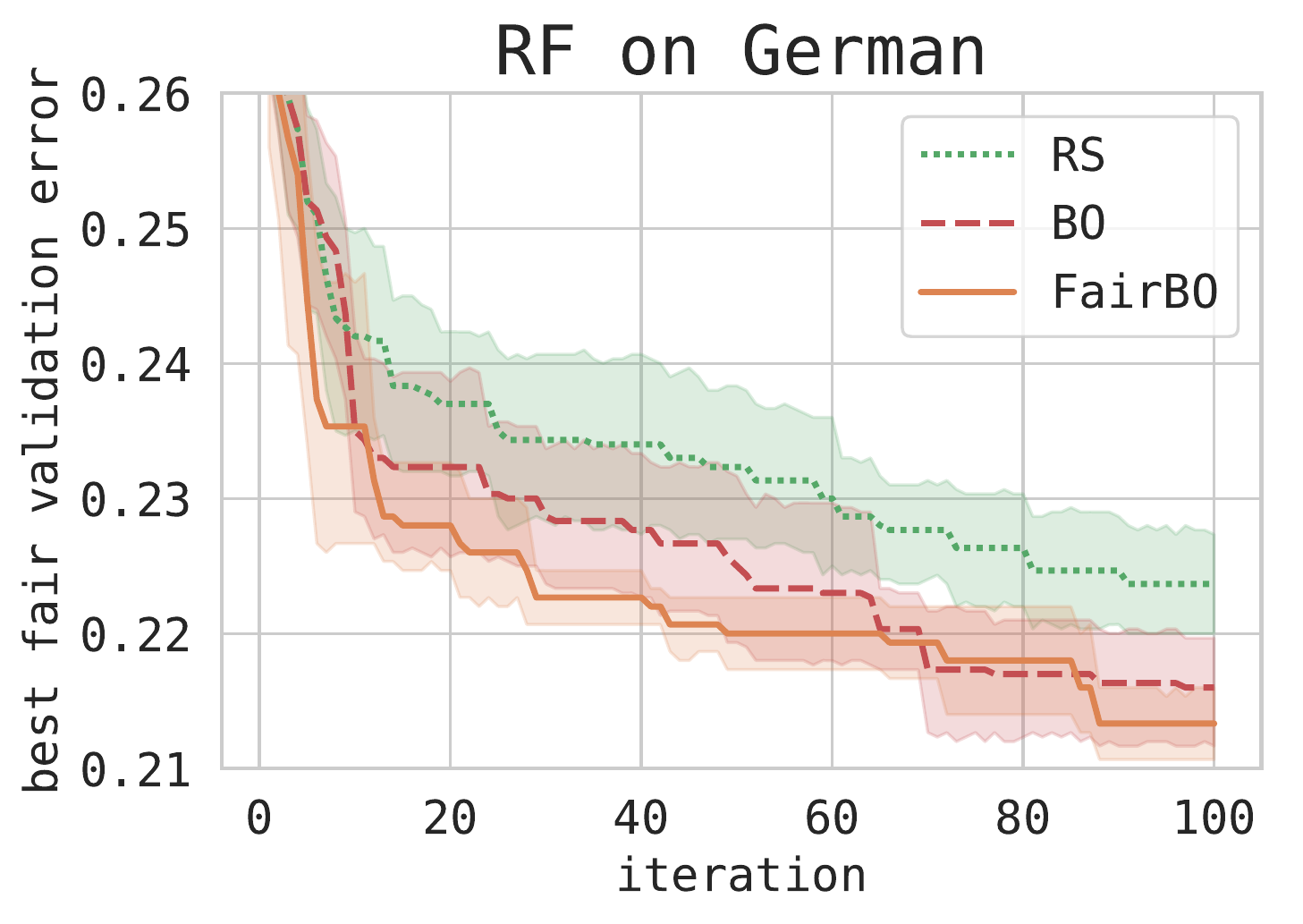}
\includegraphics[width = 0.32\textwidth]{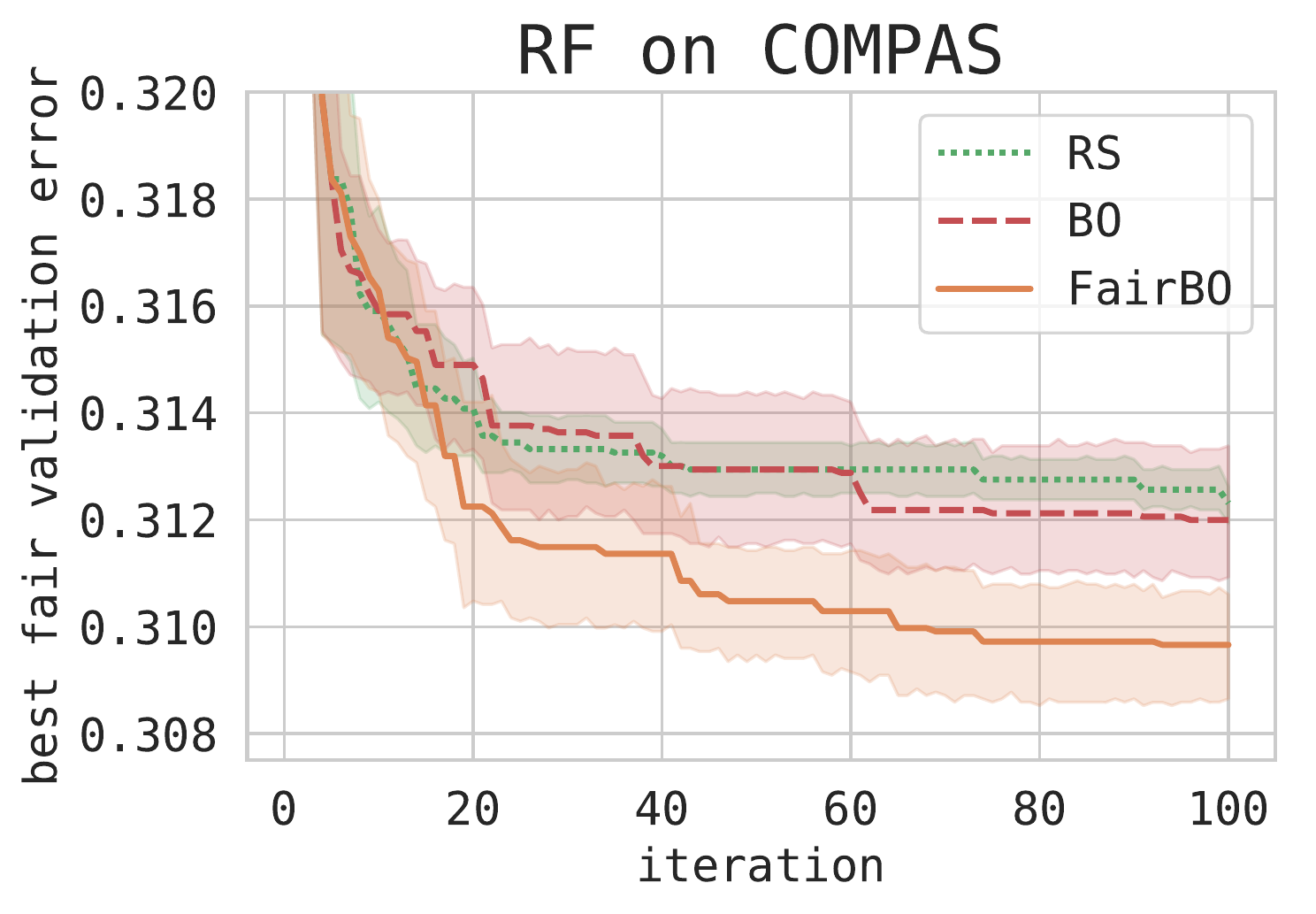}
    \caption{Comparison of RS, BO, and FairBO over the validation error (vertical axis) of the best feasible solution on Adult, German, and COMPAS data using RF as base model. The fairness constraint is set to a looser value of DSP $\leq$ 0.15. FairBO tends to find a well-performing fair model in fewer iterations than RS and BO. }
    \label{fig:bo-performance-eps15}
    \centering
    \includegraphics[width = 0.32\textwidth]{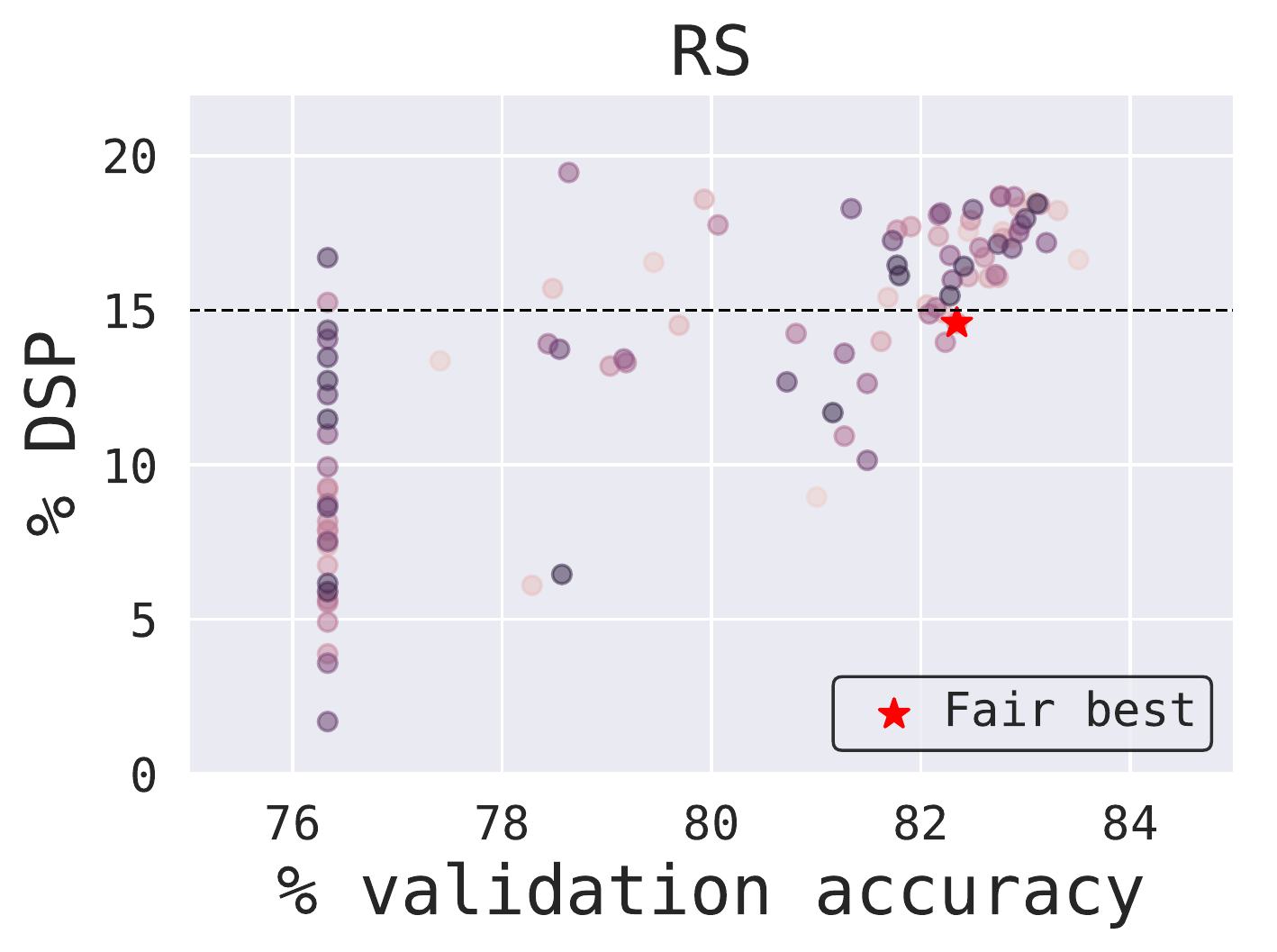}
    \includegraphics[width = 0.32\textwidth]{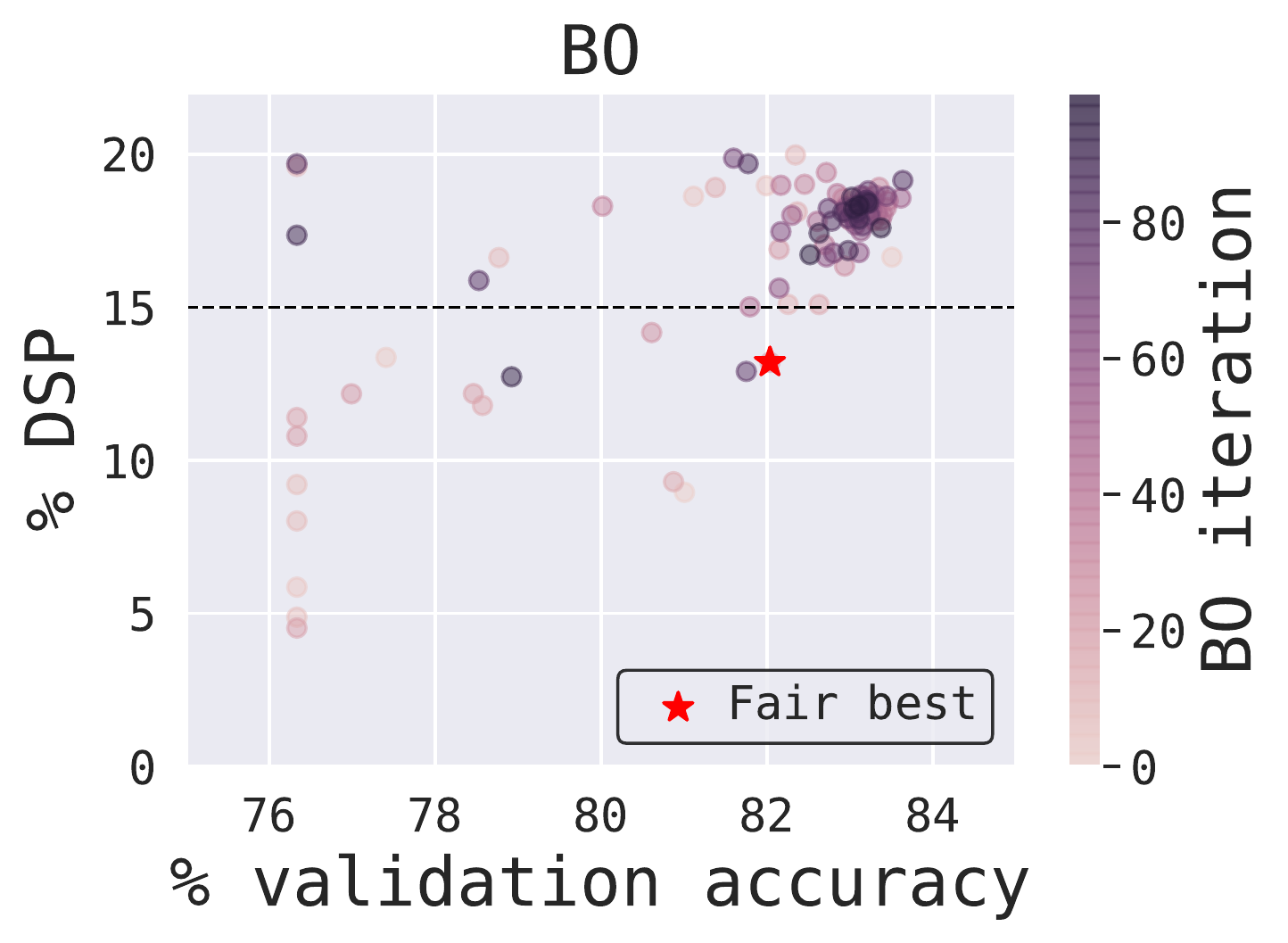}
        \includegraphics[width = 0.32\textwidth]{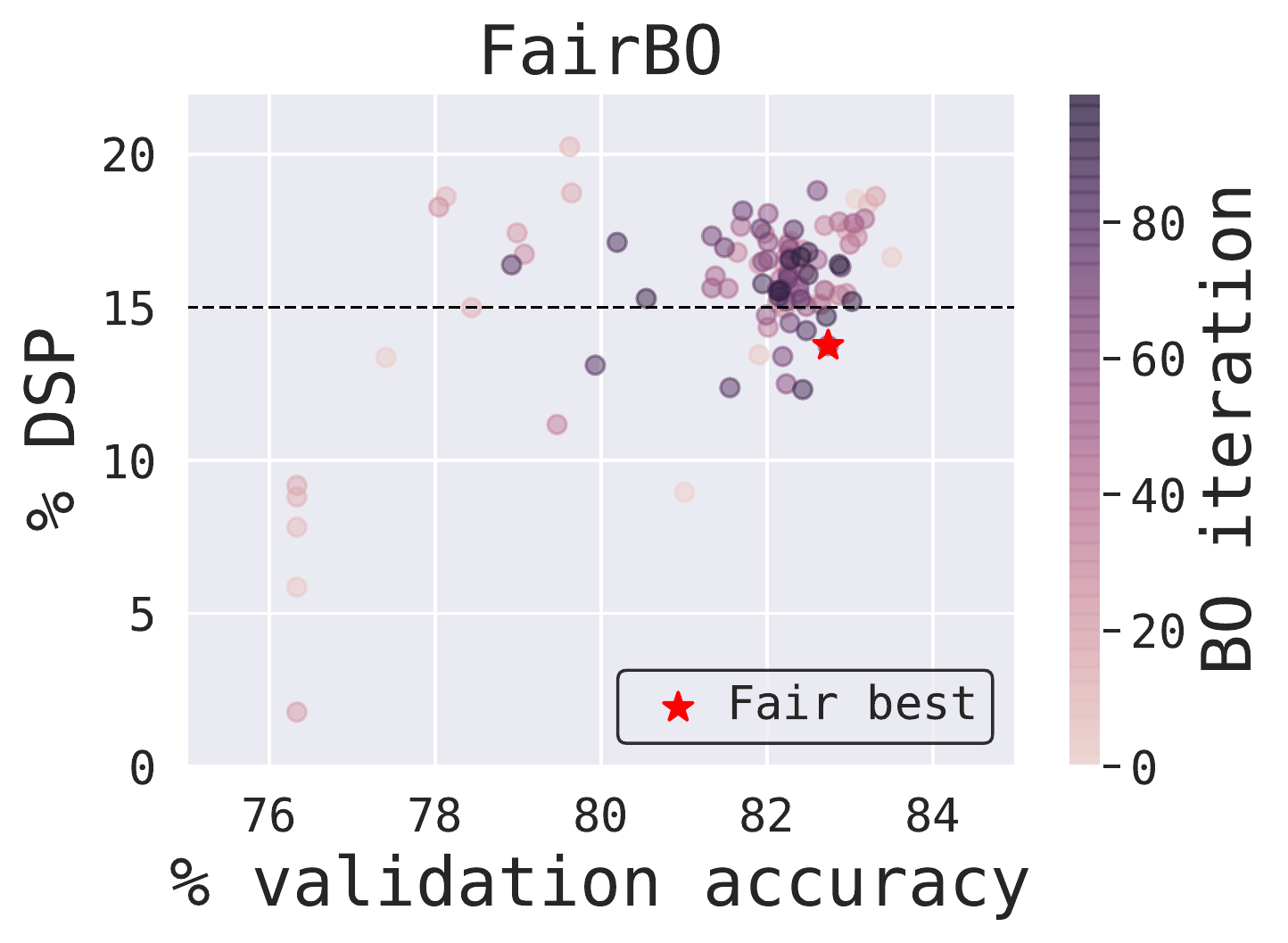}
    \caption{Comparison of FairBO, BO, and RS when tuning RF on Adult. The horizontal line is the fairness constraint, set to a looser value of DSP $\leq$ 0.15. Darker dots correspond to later BO iterations. Both RS and BO find a less accurate fair solution after 100 iterations. Additionally, BO can get stuck in high-performing yet unfair regions.}
    \label{fig:adult-results-eps15}
\end{figure*}

\paragraph{Fairness thresholds}
We investigate the impact of varying the fairness threshold $\epsilon$. We repeat the experiment of tuning a RF model on Adult, German and COMPAS, this time with a looser fairness constraint of DSP $ \leq 0.15$. Figure \ref{fig:bo-performance-eps15} shows the validation error of the fair solution on the three datasets. As expected, the performance gap between FairBO and the baselines is still clear but overall less pronounced compared to the experiments with a stricter fairness constraint DSP $\leq 0.05$. Additionally, due to the looser constraint, the accuracy of the best fair solution is significantly higher in Adult and COMPAS. At the same time, FairBO still tends to find a well-performing fair model more quickly than RS and BO. Figure~\ref{fig:adult-results-eps15} illustrates the behavior on Adult, indicating that standard BO can still get stuck in high-performing but unfair regions. Although BO also finds a fair solution, this is less accurate than the one found by FairBO. RS also tends to require more resources to find an accurate and fair solution.

\begin{figure*}[h]
    \centering
        \includegraphics[width = 0.32\textwidth]{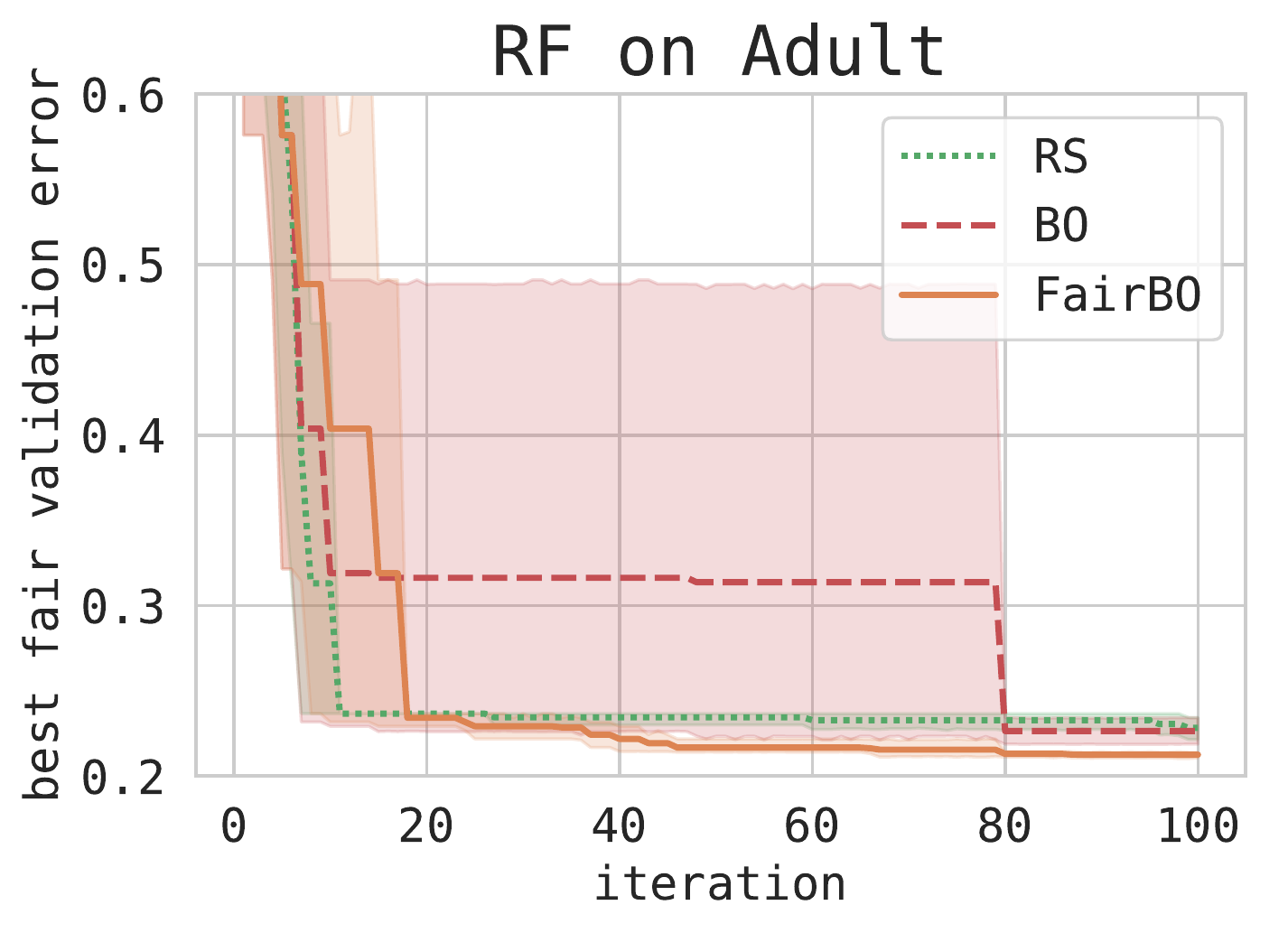}
    \includegraphics[width = 0.32\textwidth]{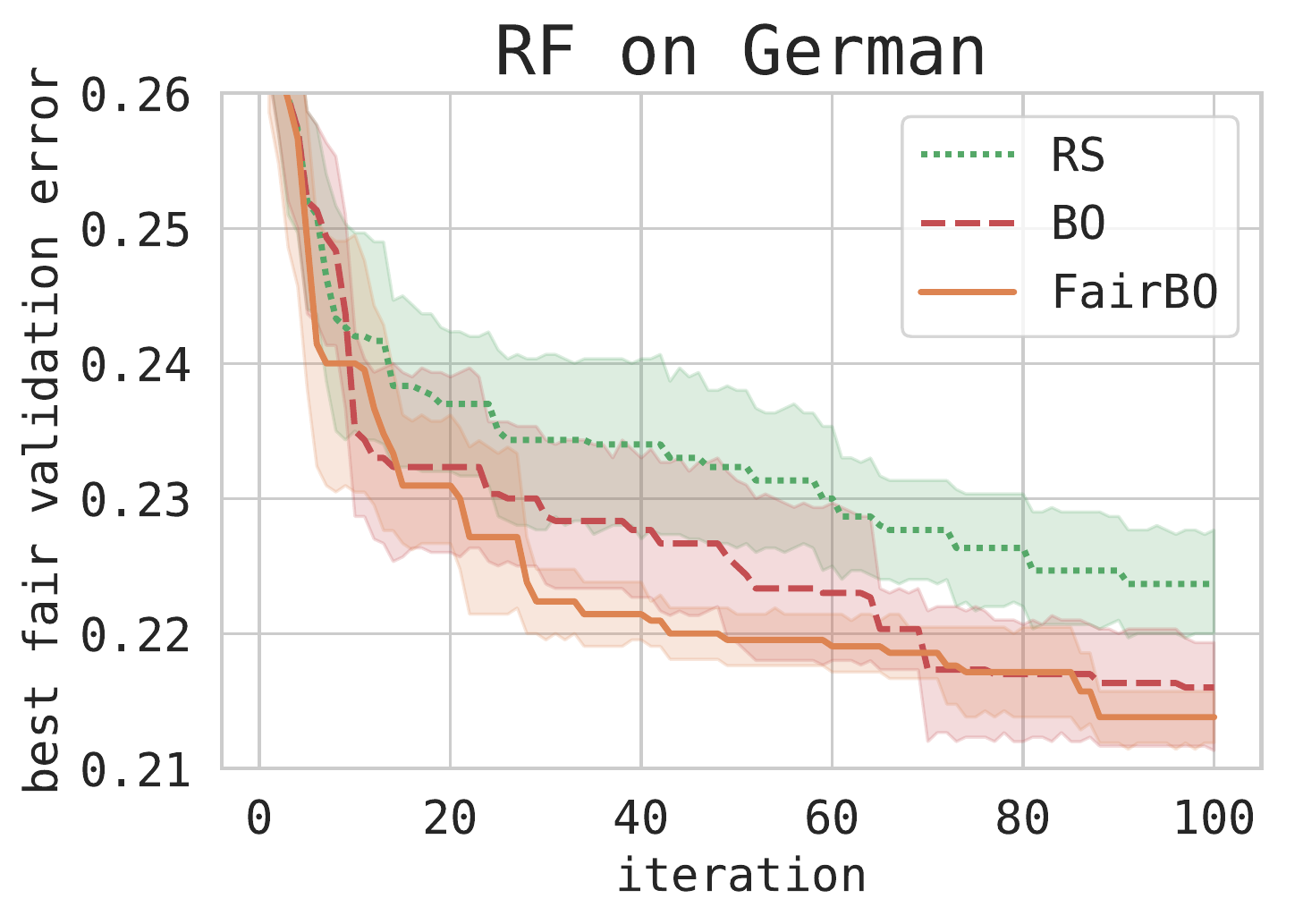}
\includegraphics[width = 0.32\textwidth]{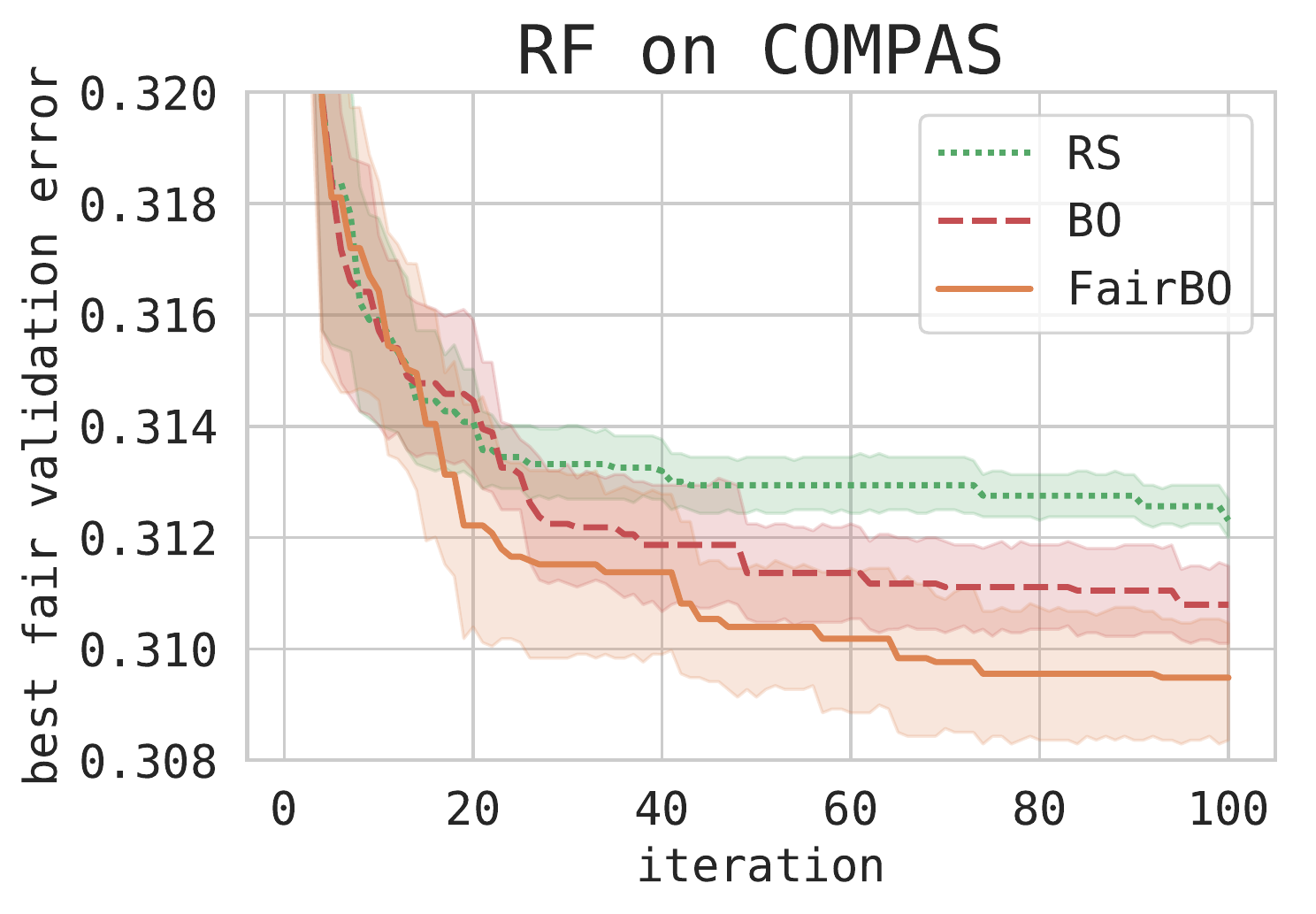}
    \caption{Comparison of RS, BO, and FairBO over the validation error (vertical axis) of the best feasible solution on Adult, German, and COMPAS data using RF as base model. The fairness constraint is set to DEO $\leq$ 0.05. FairBO tends to find a well-performing fair model more with fewer iterations than RS and BO. }
    \label{fig:adult-results-EO}
\end{figure*}

\paragraph{Fairness definitions}
FairBO can be applied directly to arbitrary fairness definitions. In this section, we repeat the RF experiments by replacing the initial, strict constraint on statistical parity (DSP $\leq$ 0.05) with an analogous constraint on equal opportunity (DEO $\leq$ 0.05). As previously, 100 iterations are allocated to RS, BO and FairBO. Figure \ref{fig:adult-results-EO} shows the best validation error of the fair solution found after each hyperparameter evaluation on the three datasets. Consistently with the DSP constraint, FairBO finds fair and accurate solutions more quickly than BO and RS. In addition, constraining DEO on COMPAS allows for fair solutions with higher accuracy compared to constraining DSP. This is not surprising as DEO is generally a better proxy of accuracy than DSP (a perfect classifier has DEO equal to zero, unlike DSP).

\begin{figure*}[h]
    \centering
    \includegraphics[width = 0.325\textwidth, trim={0.2cm 0.5cm 0.5cm 0.0cm}, clip]{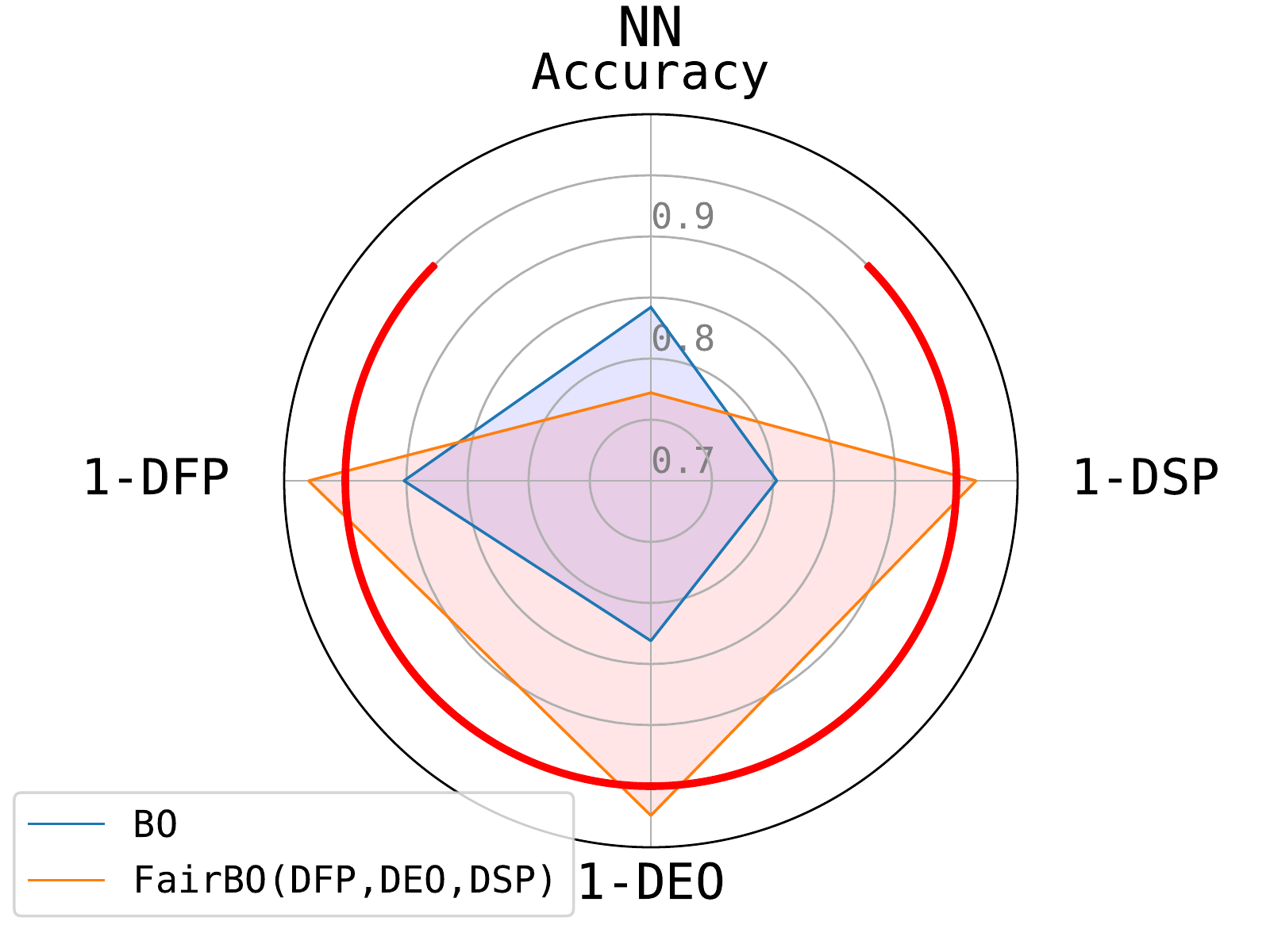}
    \includegraphics[width = 0.325\textwidth, trim={0.2cm 0.5cm 0.5cm 0.0cm}, clip]{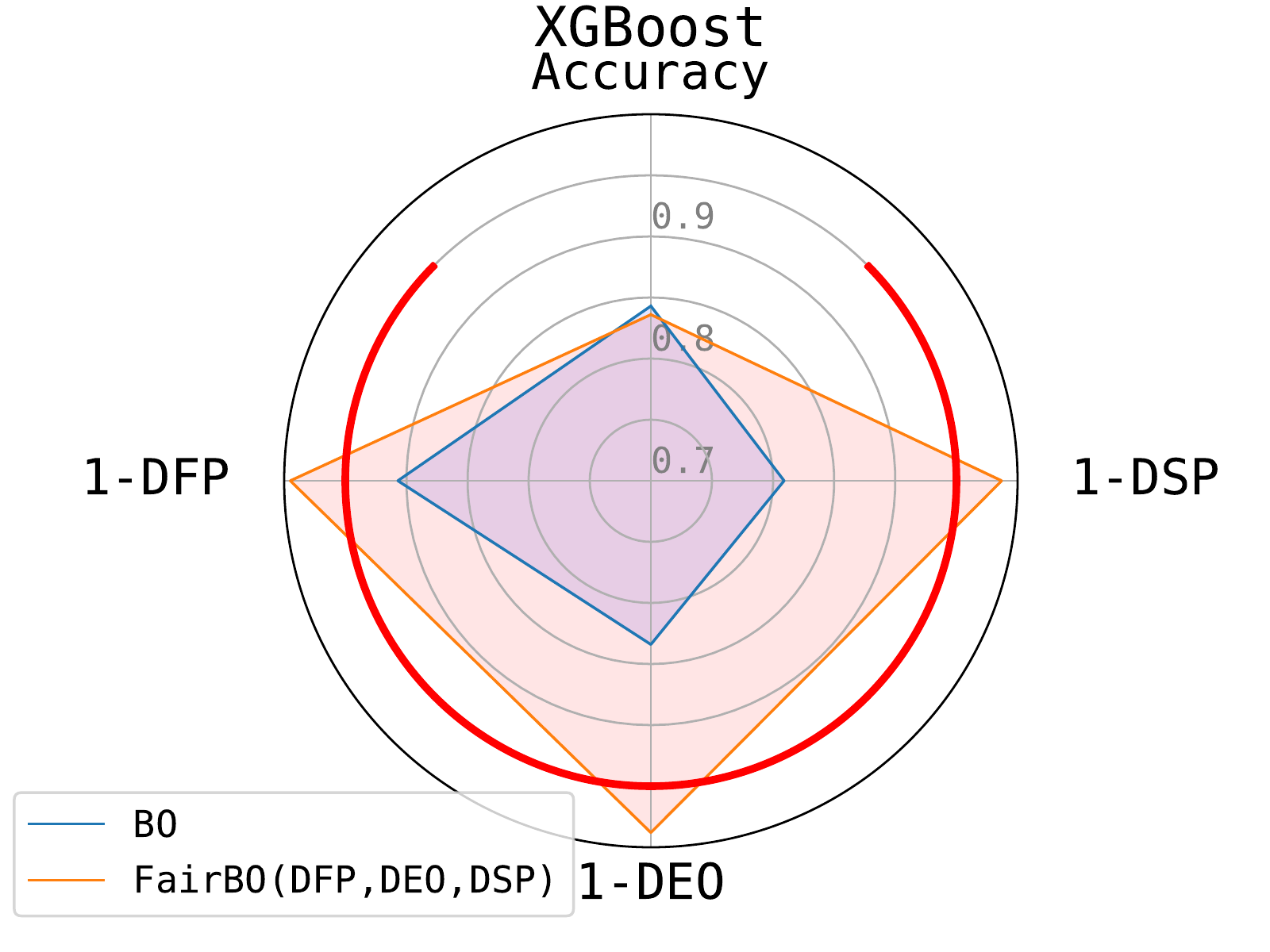}
     \includegraphics[width = 0.325\textwidth, trim={0.2cm 0.5cm 0.5cm 0.0cm}, clip]{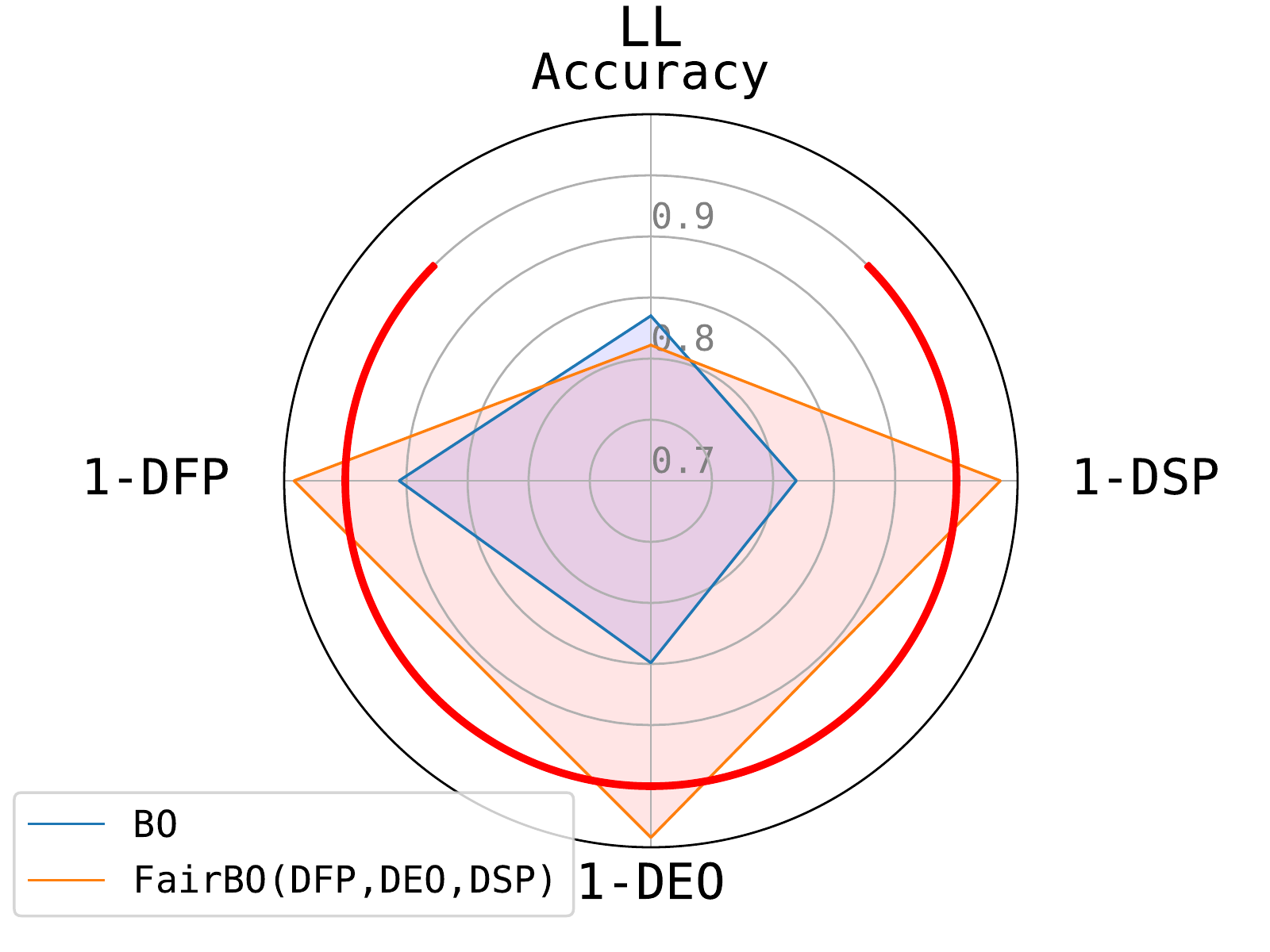}
 \caption{Best hyperparameter configuration found on Adult by BO and FairBO on NN, XGBoost, and LL with three simultaneous fairness constraints, represented by the red arches: \{DFP $< 0.05$, DEO $< 0.05$, DSP $< 0.05$\} (i.e., the minimum level of fairness we can accept). Compared to BO, FairBO can trade off accuracy for a fairer solution with respect to each definition. When applied to XGBoost, a slight accuracy degradation is enough to find solutions that are fair across all definitions.}
 \label{fig:polar2}
\end{figure*}

FairBO can also handle multiple fairness definitions simultaneously. As in the experiments with RF in the main paper, we consider 100 iterations of standard BO and FairBO on the problem of tuning NN, XGBoost and LL on Adult while satisfying the three fairness definitions. Specifically, we impose that DFP, DEO and DSP should all be less than 0.05. Results are averaged over 10 independent repetitions. Figure~\ref{fig:polar2} shows accuracy and three fairness metrics of the returned fair solutions, with the red arches indicating the constraints. FairBO allows us to trade off a relatively small degree of accuracy to get a more fair solution. Interestingly, when FairBO is applied to XGBoost, the fair solution comes with the slightest accuracy loss.

\begin{figure*}[h]
    \centering
        \includegraphics[width = 0.32\textwidth]{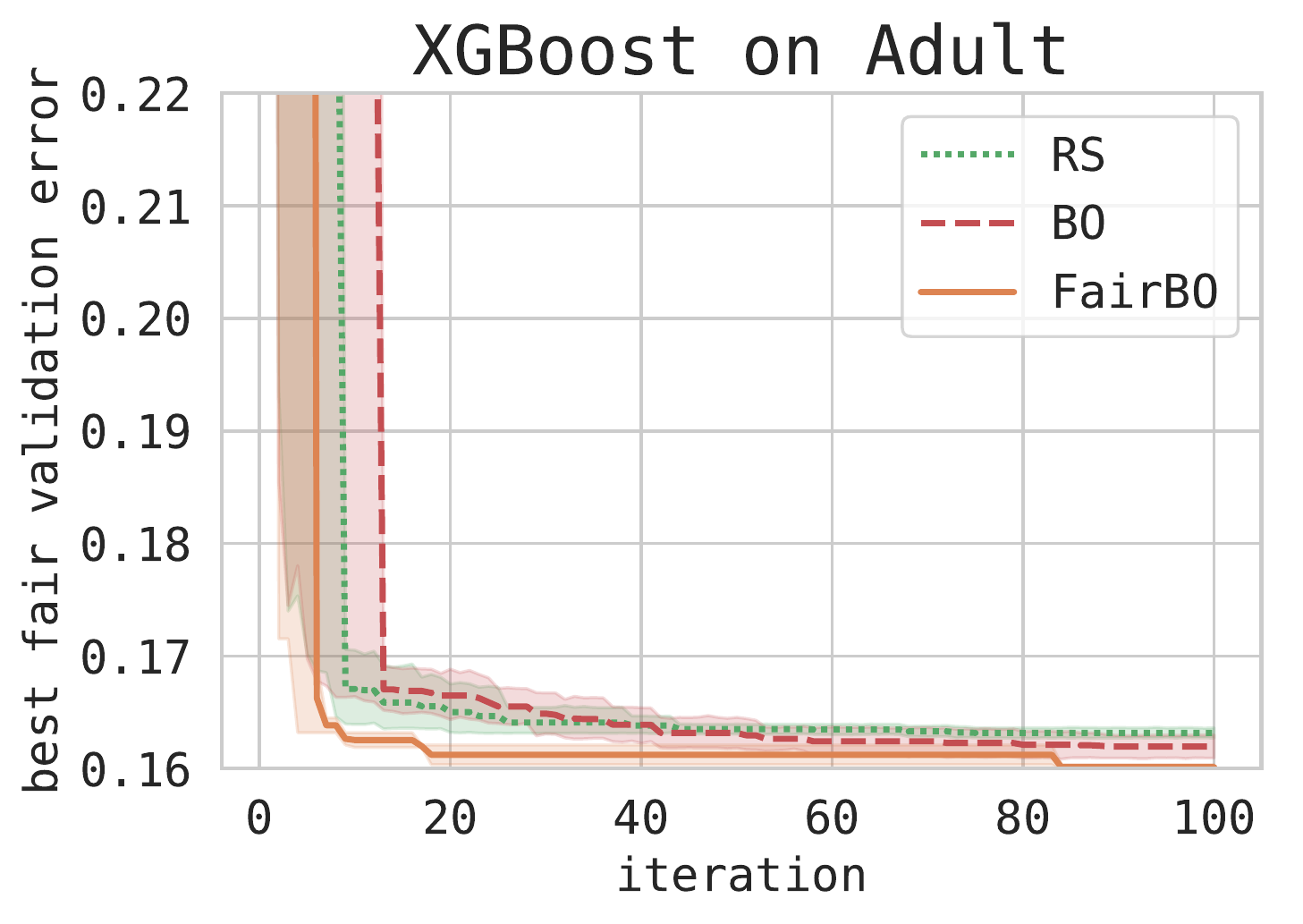}
    \includegraphics[width = 0.32\textwidth]{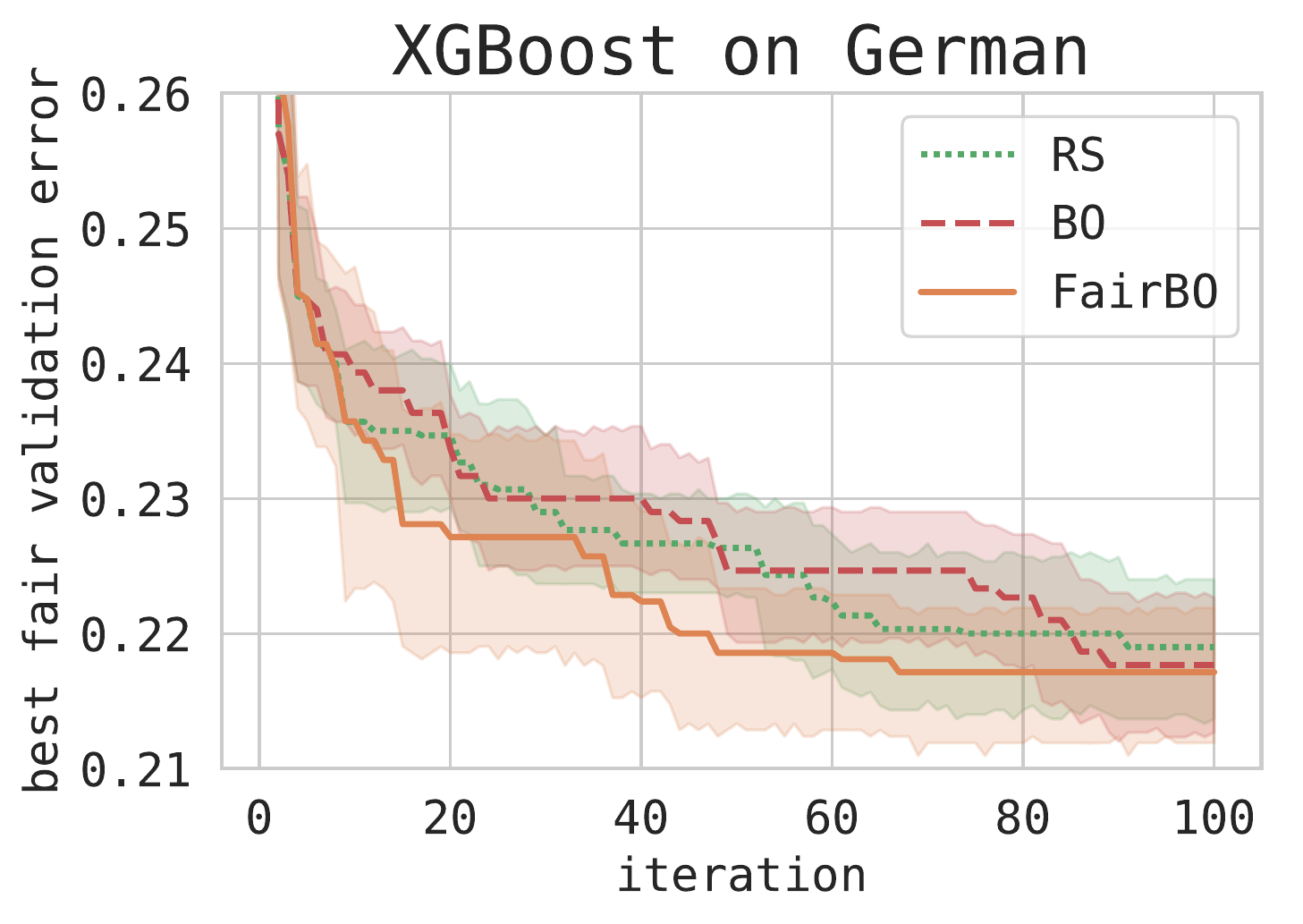}
\includegraphics[width = 0.32\textwidth]{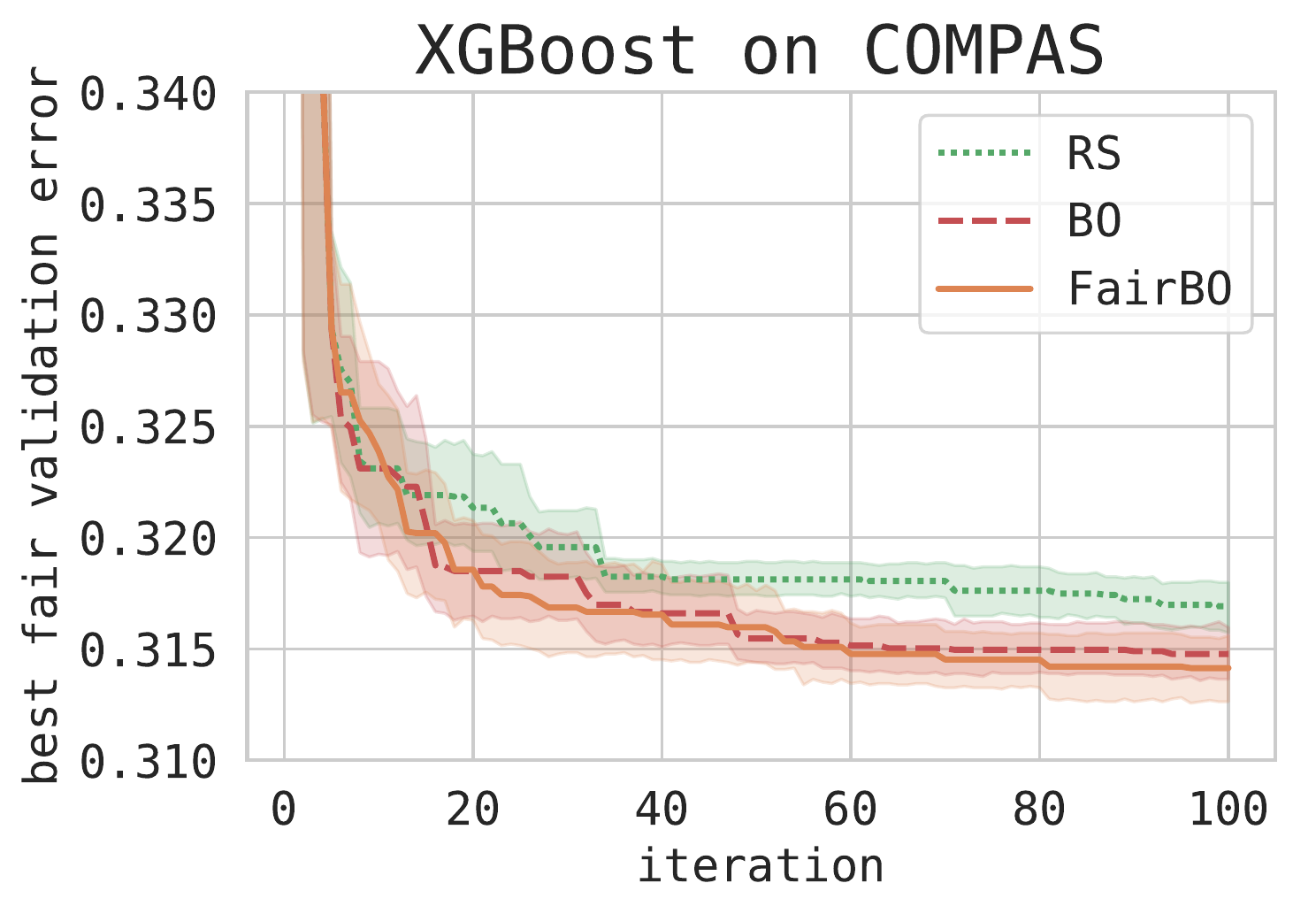}
           \includegraphics[width = 0.32\textwidth]{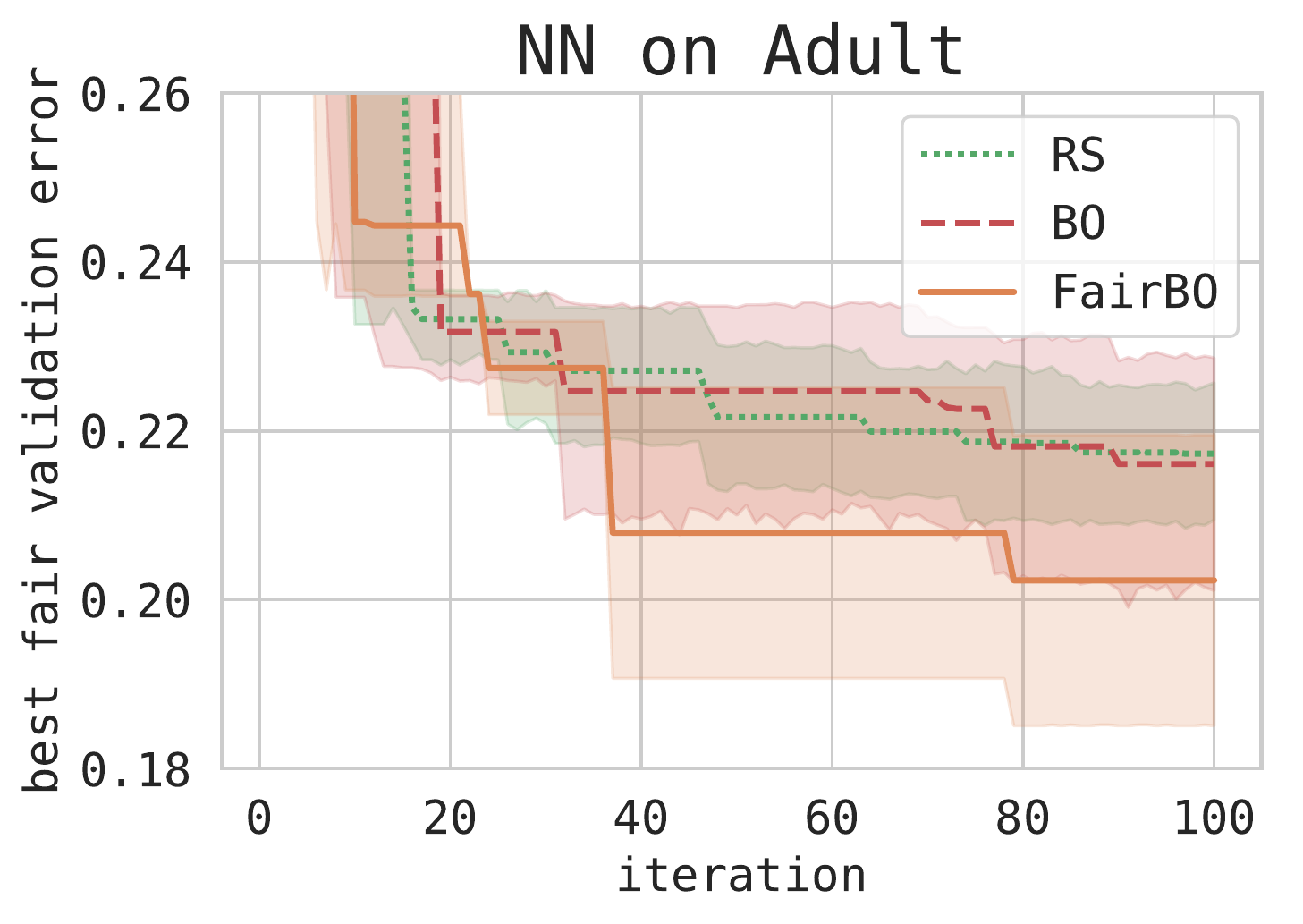}
    \includegraphics[width = 0.32\textwidth]{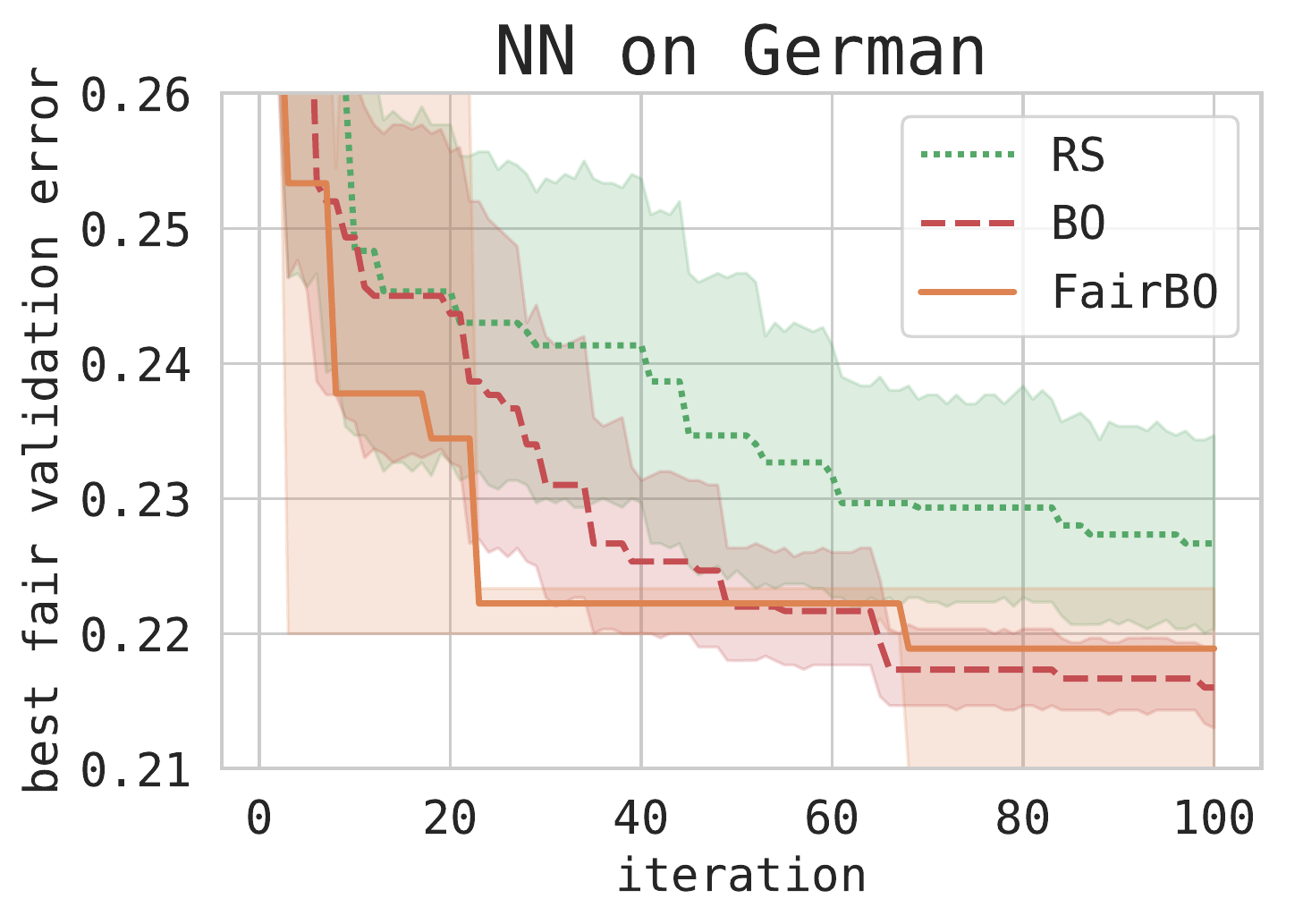}
\includegraphics[width = 0.32\textwidth]{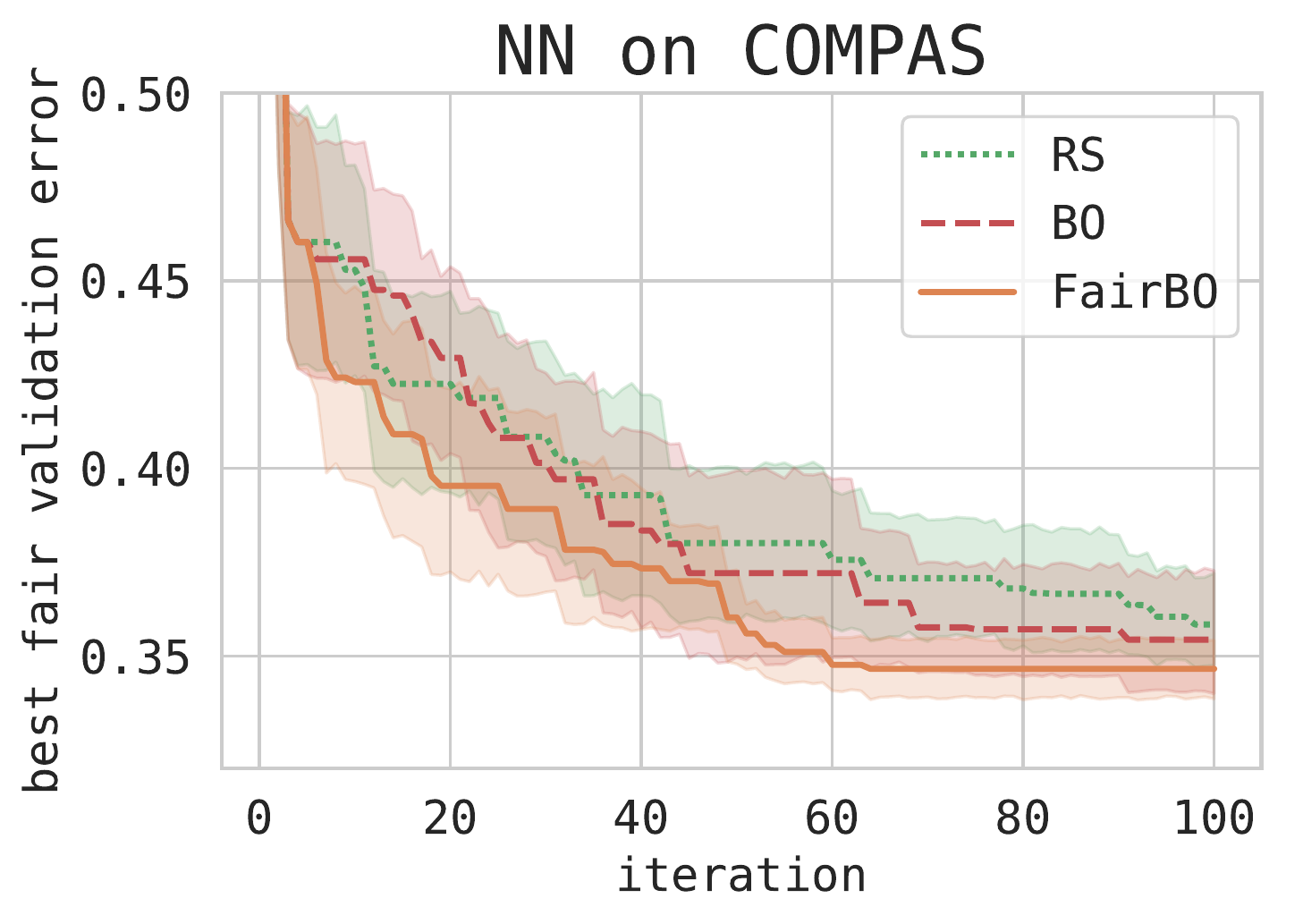}
    \caption{Comparison of RS, BO, and FairBO over the validation error (vertical axis) of the best feasible solution on Adult, German, and COMPAS data using XGBoost and NN as base models. The fairness constraint is set to DSP $\leq$ 0.05. FairBO tends to find a well-performing fair model in fewer iterations than RS and BO.}
    \label{fig:bo-performance-xgboost-nn}
\end{figure*}

\paragraph{Tuning XGboost and NN}
Figure \ref{fig:bo-performance-xgboost-nn} compares RS with BO and FairBO on the problem of tuning XGboost and a NN on Adult, COMPAS, and German. Solutions are considered fair if they satisfy the strict requirement of DSP $ \leq 0.05$. As in the case of RF, FairBO tends to find an accurate and fair solution faster than baselines. Consistently with the multiple constraint experiment, tuning XGBoost allows for more accurate fair solutions than RF and NN.

\begin{figure*}[h]
    \centering
        \includegraphics[width = 0.32\textwidth]{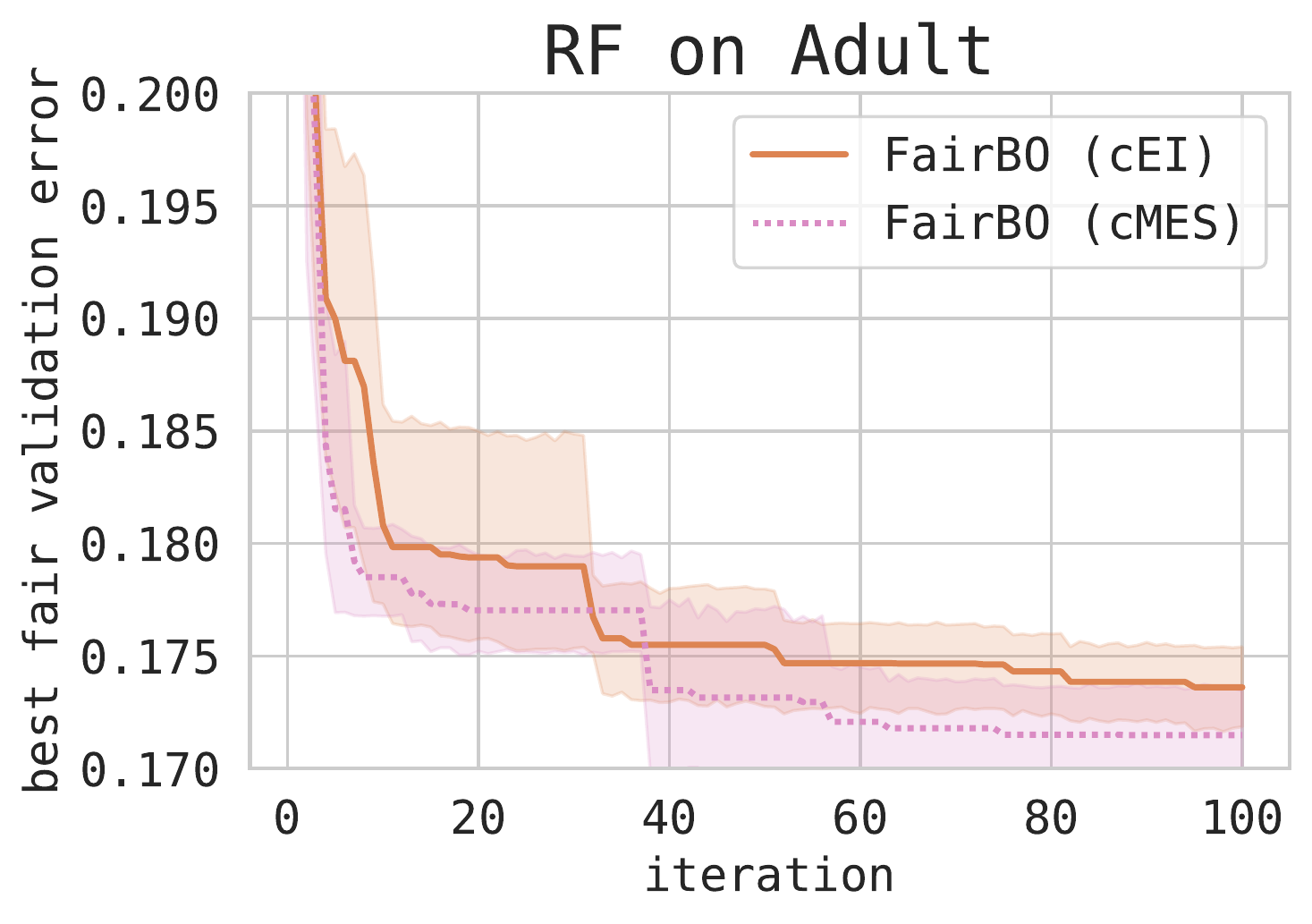}
    \includegraphics[width = 0.32\textwidth]{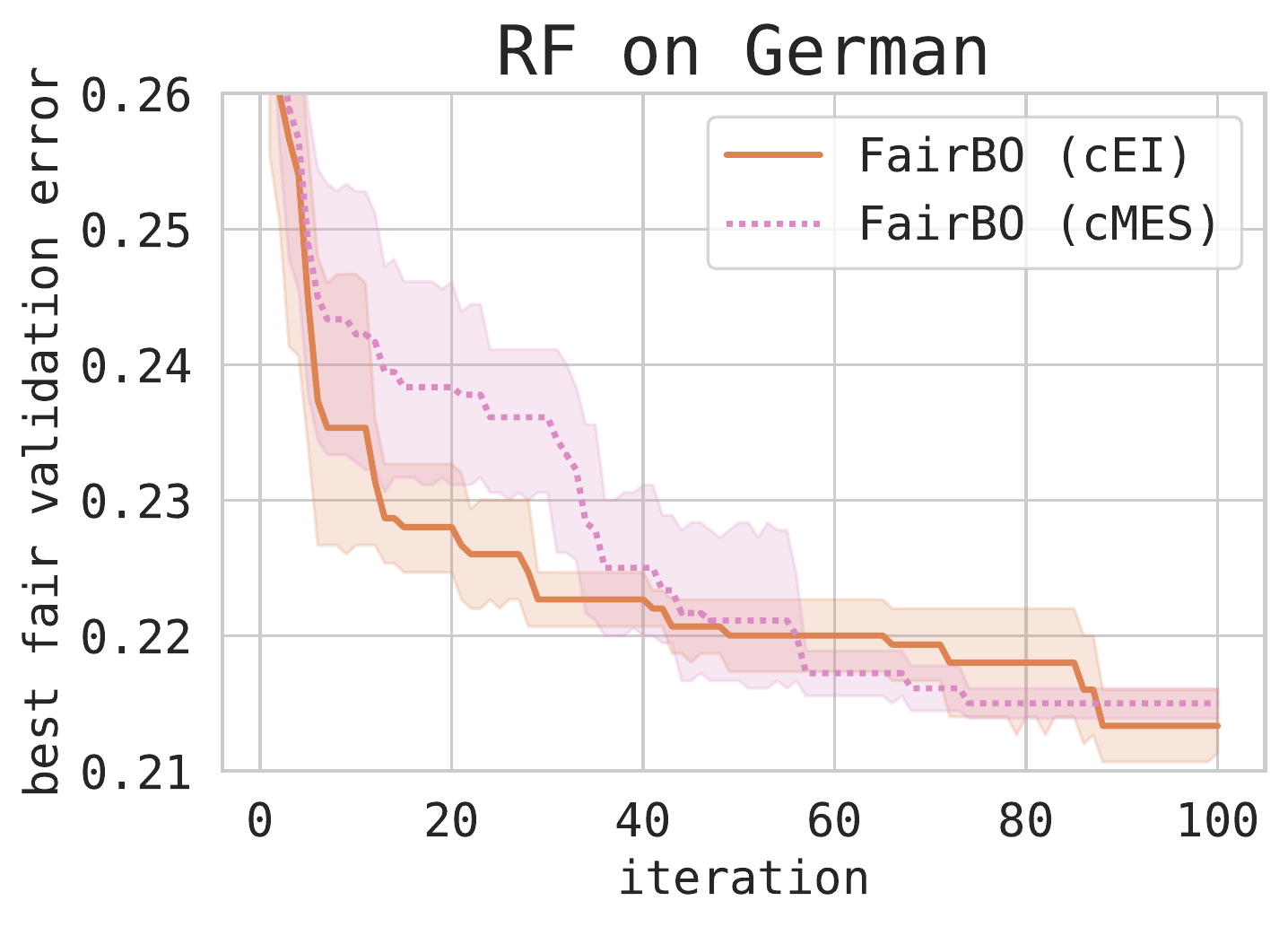}
\includegraphics[width = 0.32\textwidth]{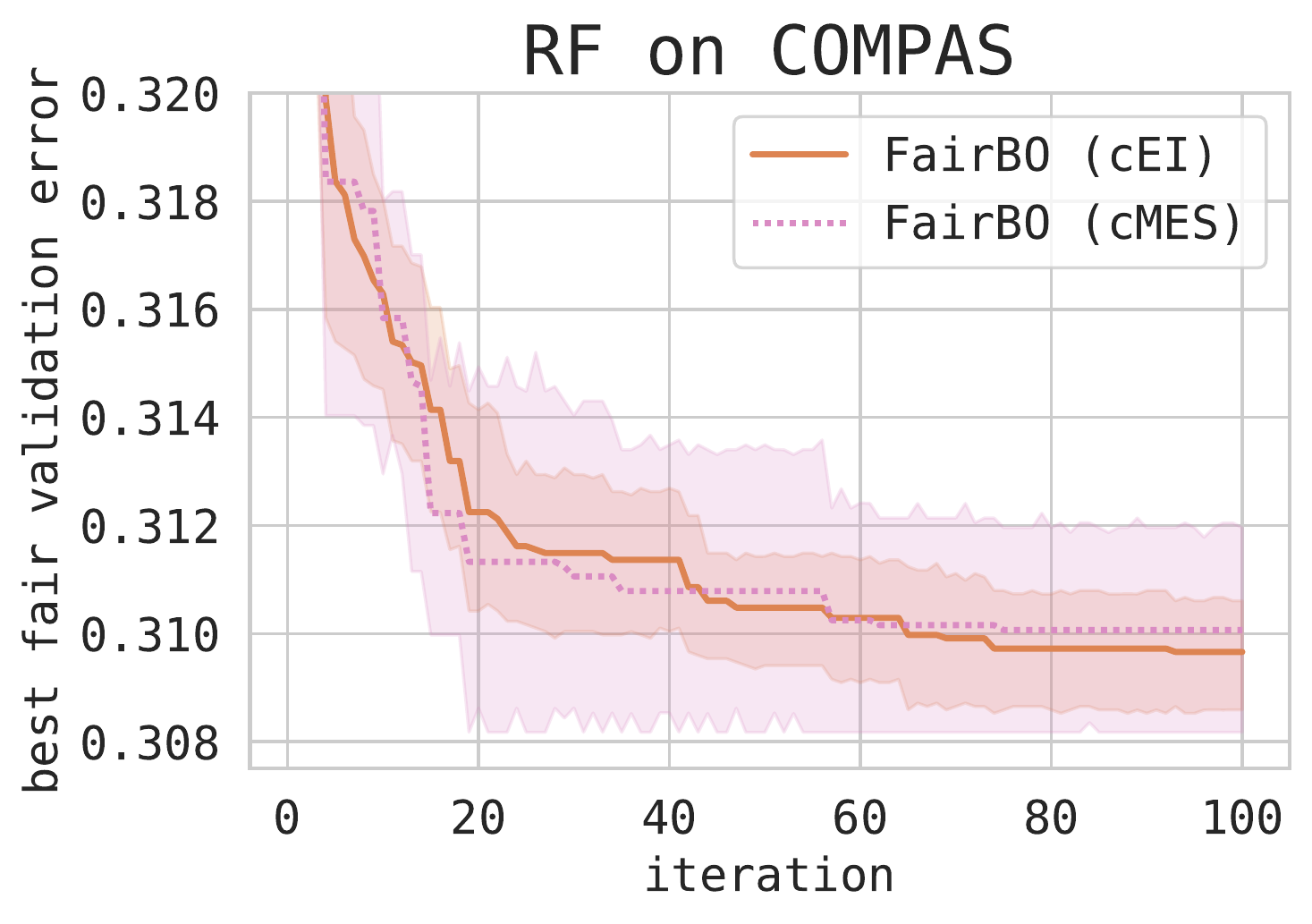}
    \caption{Comparison of cEI and cMES, two instantiations of FairBO based on the constrained expected improvement and constrained max-value entropy search acquisition function, respectively. We compare the validation error (vertical axis) of the best feasible solution on Adult, German, and COMPAS with a constraint set to DSP $\leq$ 0.15. The two instantiations of FairBO achieve comparable performance.}
    \label{fig:bo-performance-mes}
\end{figure*}

\paragraph{Acquisition functions}
Being based on a general constrained BO framework, FairBO can be implemented through different acquisition functions beyond cEI. An example is cMES, a constrained extension of the max-value entropy search [47]. Figure \ref{fig:bo-performance-mes} demonstrates this by comparing FairBO implemented through either cEI and cMES on the problem of tuning RF on Adult, COMPAS, and German. Solutions are considered fair if they satisfy the requirement of DSP $ \leq 0.15$. The results confirm that cMES is another viable implementation of FairBO which achieves consistent performance on these benchmarks. A systematic comparison of the two acquisition functions is provided in [47].

\begin{figure*}[h]
    \centering
    \includegraphics[width = 0.32\textwidth]{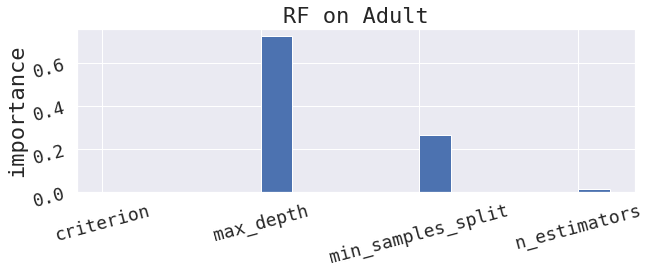}
   \includegraphics[width = 0.32\textwidth]{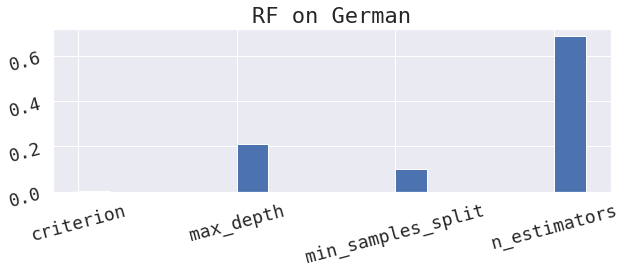}
   \includegraphics[width = 0.32\textwidth]{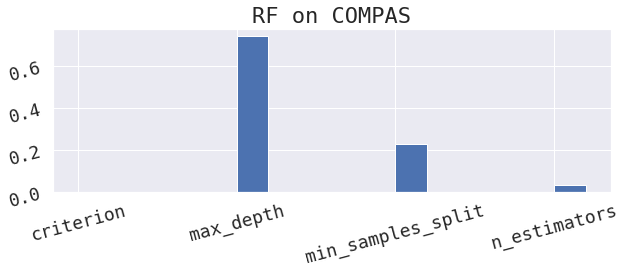}
        \includegraphics[width = 0.32\textwidth]{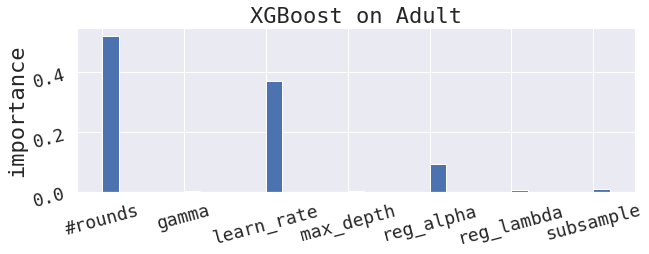}
   \includegraphics[width = 0.32\textwidth]{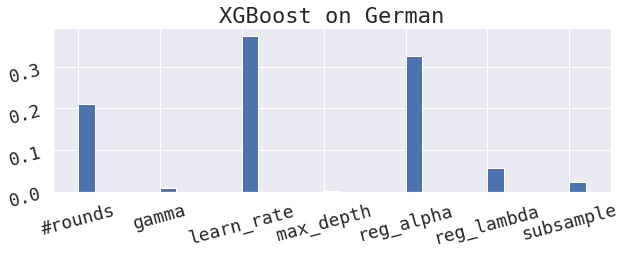}
   \includegraphics[width = 0.32\textwidth]{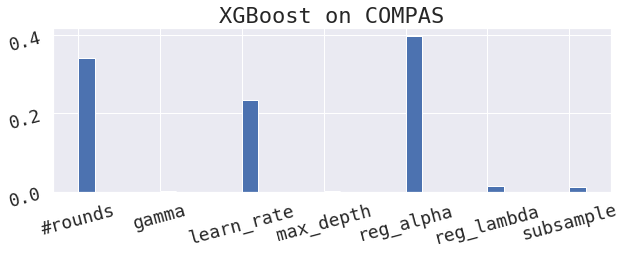}
           \includegraphics[width = 0.32\textwidth]{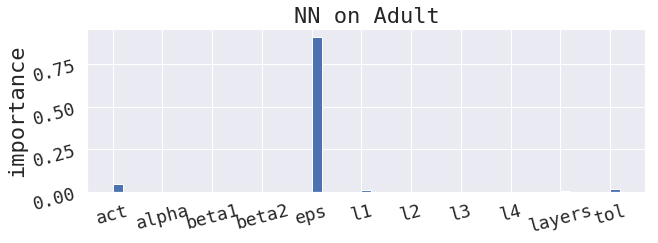}
   \includegraphics[width = 0.32\textwidth]{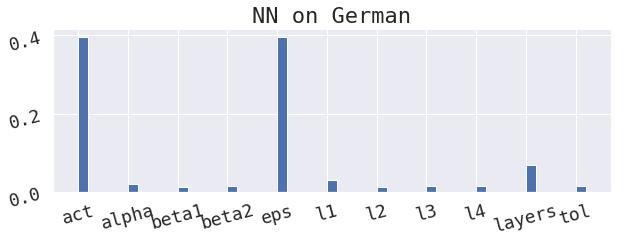}
   \includegraphics[width = 0.32\textwidth]{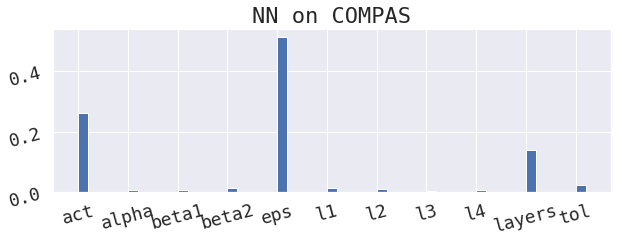}
           \includegraphics[width = 0.32\textwidth]{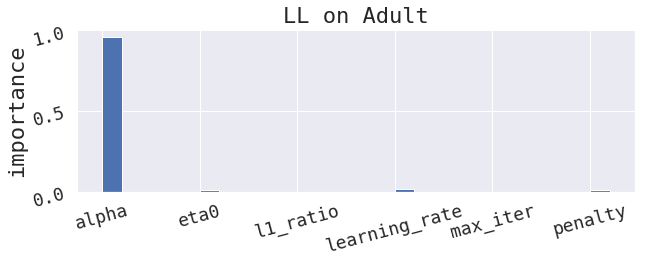}
   \includegraphics[width = 0.32\textwidth]{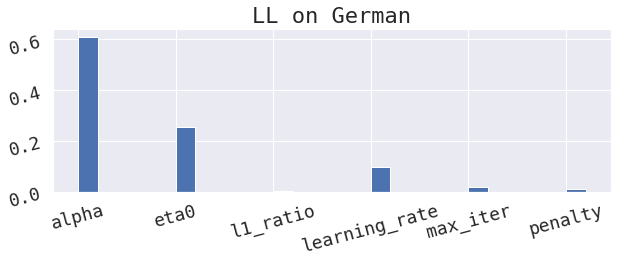}
   \includegraphics[width = 0.32\textwidth]{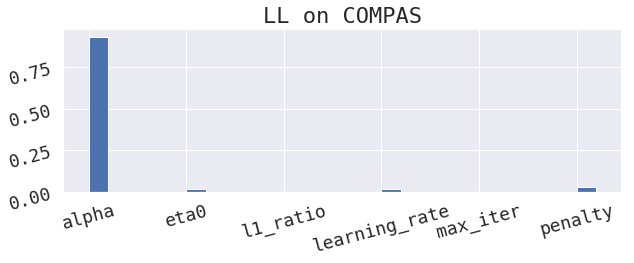}
\caption{Hyperparameter importance on equal opportunity. The role of each hyperparameter on the unfairness (DEO) of the resulting model is evaluated via fANOVA. Statistics are collected from 100 iterations of random search and 10 seeds. Regularization hyperparameters tend to impact fairness the most (e.g., LL's most important hyperparameter is precisely the regularization factor \texttt{alpha}).}
    \label{fig:fanovaEO}
\end{figure*}

\paragraph{Hyperparameters and equal opportunity}
Previous experiments shed light on the role of hyperparameter tuning on unfairness. We now study the contribution of each hyperparameter when unfairness is defined as difference in equal opportunity (DEO). As in the DSP experiments in the main paper, for each algorithm we study hyperparameter importance via fANOVA. Hyperparameter configurations and unfairness metrics are collected from 100 iterations of random search and 10 random seeds for each dataset. Figure~\ref{fig:fanovaEO} shows hyperparameter importance on DEO, confirming the results obtained with DSP and indicating that the hyperparameters controlling regularization tend to play the largest role. In the case of RF, the most important hyperparameter is the maximum tree depth on Adult and COMPAS, and number of trees on German; for XGBoost, the L1 weight regularizer \texttt{alpha}, number of boosting rounds or learning rate are the most important; for NN, the most impactful is Adam's initial learning rate \texttt{eps} (as the number of epochs is kept fixed to the default value); finally, the most relevant hyperparameter in LL is, once again, exactly the regularization factor \texttt{alpha}.

\section*{Ethical Statement}
\label{sec:impact}
Algorithmic fairness has the potential for a profound impact on society. This paper aims to make it safer to use automatic agents that generate decisions affecting critical domains such as justice, financial landing, and hiring. We believe that simplifying the process of training accurate and fair models can help spread good practice in our field and foster the generation of unbiased models. Our work pursues exactly this goal. 

More fair machine learning is needed in our society, especially in light of the many discoveries of negative bias in commonly used machine learning models. With less biased and more fair machine learning, we can improve the trust of our automatic agents and increase awareness on this topic among the colleagues in our community. But to maximize impact, fair machine learning also needs to be accessible to non-experts. Ultimately, we have the possibility to enhance the benefits that machine learning can bring to society without translating human biases to the learned models.

We are aware that statistical measures of fairness, such as statistical parity or equal opportunity, cannot be considered as the unique definitions. Indeed, any definition of fairness applied to the task at hand has to be carefully understood and chosen by a human, not by an automatic agent. It is well-known that some of the definitions are in contrast with each other so that, by enforcing one, we are simultaneously forcing other definitions to be violated. The choice of the right definition is fundamental but out of the scope of our proposal, and requires a human-in-the-loop approach.